\documentclass[sigconf,nonacm=true]{acmart}
\usepackage[color=green!40]{todonotes}
\usepackage{graphicx}
\usepackage{subcaption}
\usepackage{xcolor}

\AtBeginDocument{%
  }

\acmISBN{978-1-4503-XXXX-X/2018/06}

\begin{document}

\title{VirtualFencer: Generating Fencing Bouts based on Strategies Extracted from In-the-Wild Videos}

\author{Zhiyin Lin}
\email{zhiyinl@stanford.edu}

\affiliation{%
  \institution{Stanford University}
  \country{USA}
}

\author{Purvi Goel}
\email{pgoel2@cs.stanford.edu}

\affiliation{
  \institution{Stanford University}
  \country{USA}
}
\author{Joy Yun}
\email{joyyun@stanford.edu}

\affiliation{%
  \institution{Stanford University}
  \country{USA}
}
\author{C. Karen Liu}
\email{karenliu@cs.stanford.edu}

\affiliation{%
  \institution{Stanford University}
  \country{USA}
}
\author{Joao Pedro Araujo}
\email{jparaujo@stanford.edu}

\affiliation{%
  \institution{Stanford University}
  \country{USA}
}

\begin{abstract}

Fencing is a sport where athletes engage in diverse yet strategically logical motions. While most motions fall into a few high-level actions (e.g. step, lunge, parry), the execution can vary widely—fast vs. slow, large vs. small, offensive vs. defensive. Moreover, a fencer’s actions are informed by a strategy that often comes in response to the opponent’s behavior. This combination of motion diversity with underlying two-player strategy motivates the application of data-driven modeling to fencing. We present VirtualFencer, a system capable of extracting 3D fencing motion and strategy from in-the-wild video without supervision, and then using that extracted knowledge to generate realistic fencing behavior. We demonstrate the versatile capabilities of our system by having it (i) fence against itself (self-play), (ii) fence against a real fencer’s motion from online video, and (iii) fence interactively against a professional fencer.

\end{abstract}

\ccsdesc[300]{Computing methodologies~Image processing}
\ccsdesc[500]{Computing methodologies~Motion processing}

\keywords{Human Simulation, Animation, Motion Estimation \& Tracking, User Studies}
\begin{teaserfigure}
  \includegraphics[width=\textwidth]{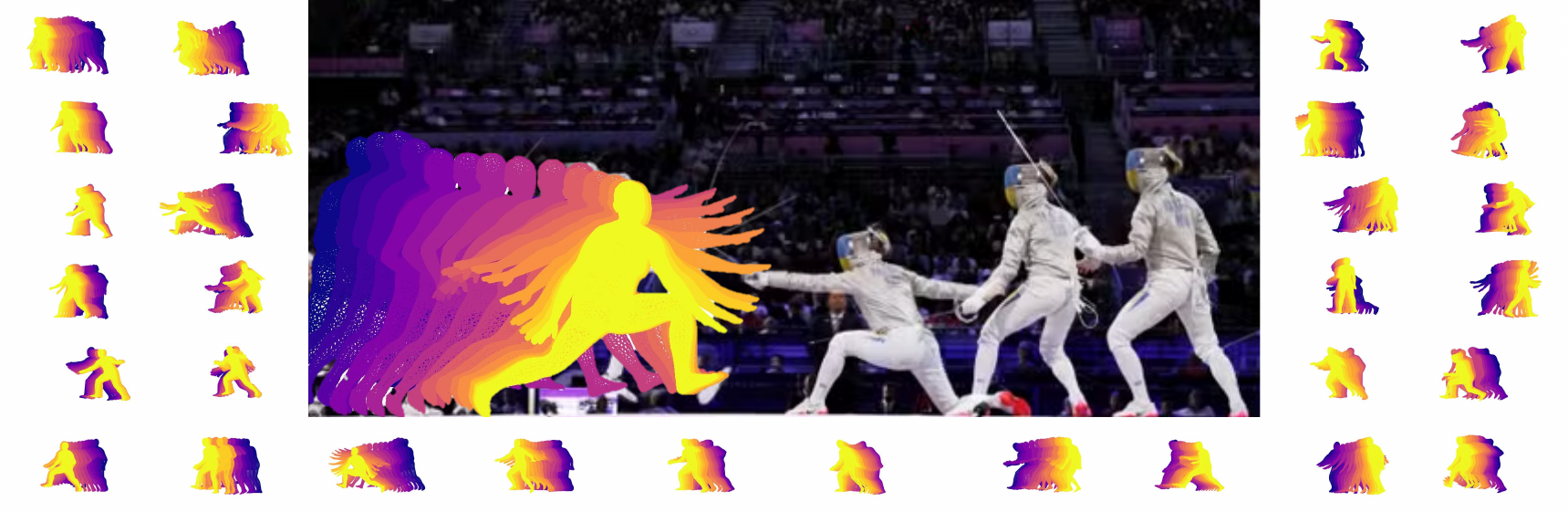}
  \caption{Our system VirtualFencer leverages low-level skills discovered unsupervisedly from in-the-wild videos and fences against (i) itself, (ii) a real fencer's motion from online video, and (iii) a professional fencer interactively.}
  \label{fig:teaser}
\end{teaserfigure}

\received{20 February 2007}
\received[revised]{12 March 2009}
\received[accepted]{5 June 2009}

\maketitle

\section{Introduction}


Athletes at every level encounter obstacles to effective training. Team sports demand a full roster, while even one-on-one sports still require specialized equipment, access to coaching, and, above all, a practice partner. Computational tools and technologies such as virtual and mixed reality hold the potential to remove some of these barriers and make sports more accessible, for example by providing digital counterparts of real-world training experiences or delivering personalized feedback \cite{Malawski2022}. 
However, materializing this potential also has associated challenges. A tool can only provide meaningful feedback if it has a good understanding of the game being played and knows what to watch for. Similarly, real-world training experiences can only be provided if they can be captured and replayed using these computational tools.


In this paper, we focus on how computational tools may help the study of fencing. Fencing is like playing physical chess, with a rich library of moves and possible strategies. The continuous nature of motion, together with the natural variability of skill execution, make data-driven analysis tools a suitable candidate for its study. However, there are several challenges that come with analyzing and modeling fencing touches. First, the taxonomy of fencing moves is complex, requiring extensive domain knowledge to understand the underlying strategic narrative that ties together individual actions. Second, large-scale datasets of high-quality annotated fencing motions do not exist, and creating one would require considerable effort. 

On the other hand, a large amount of fencing videos from competitions and tournaments is available online. We address the lack of fencing motion data by leveraging recent advances in human motion reconstruction from in-the-wild video. With the availability of fencing motion, unsupervised learning techniques (such as clustering) can be used to recover some of the structure of the fencing moves without requiring expert annotation, and then validated using only a minimal amount of manual labeling effort. Finally, the recovered structure can be exploited to generate novel fencing matches. To this end, we present VirtualFencer, a realistic fencing opponent that users can interact with.


We make three contributions in this paper:

\begin{enumerate}
    \item A system for extracting fencing motion from in-the-wild video. We use our system to extract around 1.5 hours of fencing actions from 40 bouts of fencing, featuring 54 fencers on the senior international circuit. 
    \item A method for unsupervised skill discovery from motion data. We cluster our motion data into these discovered skills and use it to model the strategies fencers use. Using this strategy model, we generate novel fencing bouts.
    \item A user study evaluating the fencing bouts generated by our method. We ask over 30 professional fencers to rank the touches generated by our method relative to the ground truth touches and touches generated at random. We find that professional fencers split their preferences evenly between touches generated by our agent and ground truth touches, validating the correctness of the strategy extracted by our tool. 
\end{enumerate}

\section{Related Work}



\subsection{Study of fencing}

Prior work studied fencing action classification. \citet{MALAWSKI20181} introduced the Fencing Footwork Dataset (FFD), a dataset of 10 fencers doing 6 skills. The motion data was collected using IMU measurements and a Microsoft Kinect. The authors propose a support vector machine based classifier to analyze fencing motion. Other works \cite{MALAWSKI2019198, Zhu_2022_CVPR} have relied on this dataset. \citet{MALAWSKI2019198} improved on their prior results by using better features for classification. \citet{Zhu_2022_CVPR} introduce FenceNet and BiFenceNet, two neural network based approaches that match prior performance on FFD, but are able to work directly from 2D skeleton keypoints. In addition to capturing the fencer's motion, \citet{malawski2018real} also looked into capturing the blade's motion .

Other works study the kinematics \cite{bober2016original} and biomechanical aspects \cite{moore2015novel} of fencing movements.

The number of possible fencing motions extends well beyond 6, thus limiting what can be achieved from FFD. Moreover, the IMU based motion capture method is harder to scale than using video data (which is available in abundance). This motivates the use of recent advances in motion reconstruction from in-the-wild video, as well as the use of unsupervised learning techniques to recover the different action types.

\subsection{Human motion from in-the-wild video}

The abundance of video content on the Internet has motivated a large body of work in extracting motion data from these videos. The design space for these methods is broad. There are methods that reconstruct the motion on a per-frame basis \cite{insafutdinov2017arttrack, goel2023humans, li2023hybrik, li2023niki, pymafx2023, SMPL-X:2019}, and others that consider the entire video \cite{humanMotionKZFM19, kocabas2019vibe, Luo_2020_ACCV, choi2020beyond, WeiLin2022mpsnet, wham:cvpr:2024}. An approach that is common to several of these methods is to start by using an off-the-shelf 2D skeleton keypoint detector \cite{8099626, Insafutdinov2016, 8099977} to initialize the human pose, and then lift this 2D information to 3D, possibly leveraging some prior learned from 3D data \cite{AMASS:ICCV:2019}. Other methods, such as the ones by \citet{hmrKanazawa17} and \citet{taylor2012vitruvian}, go directly from image to 3D mesh. Some methods assume there is only a single subject in the scene \cite{song2017thin}, while others are designed to/capable of handle multiple subjects \cite{insafutdinov2017arttrack, iqbal2017posetrack, he2017mask, girdhar2018detect, guler2018densepose}.

Finally, some methods assume a static camera, while others take camera motion into account \cite{ye2023slahmr, Kocabas2023pace, glopro, wham:cvpr:2024}.

Fencing videos allow us to make several assumptions. First, most of the action will be focused on the two fencers. Second, the motion of the two fencers will be constrained to the fencing strip (which provides us scale information). Third, the camera will remain mostly fixed, only rotating. These allow us to recover the fencers' motion in global coordinates with reasonable accuracy.

\subsection{Leveraging data for sports applications}

\citet{vid2player} leverage annotated broadcast video of tennis matches to generate interactively controllable video sprites of professional tennis players. Follow-up work \cite{zhang2023vid2player3d} lifts this to fully controllable 3D characters, and is able to handle two-player interactions.

In this work, we similarly leverage video of fencing bouts. However, since our main focus is on the strategy extraction rather than motion generation, we are more interested in collecting large amounts of motion data and studying the relationships between those motion clips. Two classical papers that are closely related to our approach are motion graphs \cite{motiongraphs} and the work by \citet{lee2002interactive}, which also uses clustering of unstructured motion data, but with the goal of being able to quickly identify motion transitions, rather than low-level skills.

\section{Background: Structure of a Fencing Bout}

The Olympic sport of fencing has three primary disciplines: épée, foil, and saber. The three disciplines differ in the type of blade that is used and the rules of play that the athletes follow. For the rest of the paper, when we refer to fencing, we are specifically discussing saber fencing. 

Many of the design decisions in our system were informed by domain-specific knowledge of saber fencing. This section will cover the high-level rules and strategies in saber fencing.

The objective of each point in fencing is to land a valid hit on the opponent's target area, known as a \textit{touch}. Within a touch, a fencer's strategy and decision-making is heavily influenced by whether they have \textit{priority}, also known as right of way. Priority determines which fencer gets the point in the case that both fencers land a hit simultaneously; having or not having priority can noticeably change the strategy that a fencer chooses.

Each touch restarts in a neutral position, known as \textit{en garde}, where neither fencer has priority. Depending on the first action of each touch, one fencer claims priority, dictating which side goes on offense and which on defense. During a touch, fencers make a series of critical decisions heavily influenced by which of the following states they are in:
\begin{enumerate}
    \item \textbf{Neutral (M-M):} When neither fencer has priority, the primary objective is to claim it. In a neutral state, priority is awarded to the side that starts their attack first, measured by factors such as speed, timing, and intention.
    \item \textbf{Offense (P-NP):} The fencer that has priority is automatically on offense, with the sole intention of landing an attack on their opponent. Given they will lose priority if they do not attack, they will advance until they are within distance to score.
    \item \textbf{Defense (NP-P):} The fencer who does not have priority is on defense. Their primary objective is to reclaim priority by tricking the opponent into making a mistake, primarily through causing the opponent to miss or by blocking their cut. When the attacker makes a mistake, priority automatically transfers to the other fencer.
\end{enumerate}

A single touch may involve multiple exchanges of priority between the fencers, yielding an intricate web of strategic decisions. Our system is capable of parsing and learning these complex dynamics from in-the-wild video.  

\section{System Overview}

\begin{figure*}[!t]
  \centering
  \includegraphics[clip,trim=0cm 1cm 0cm 1cm,width=\textwidth]{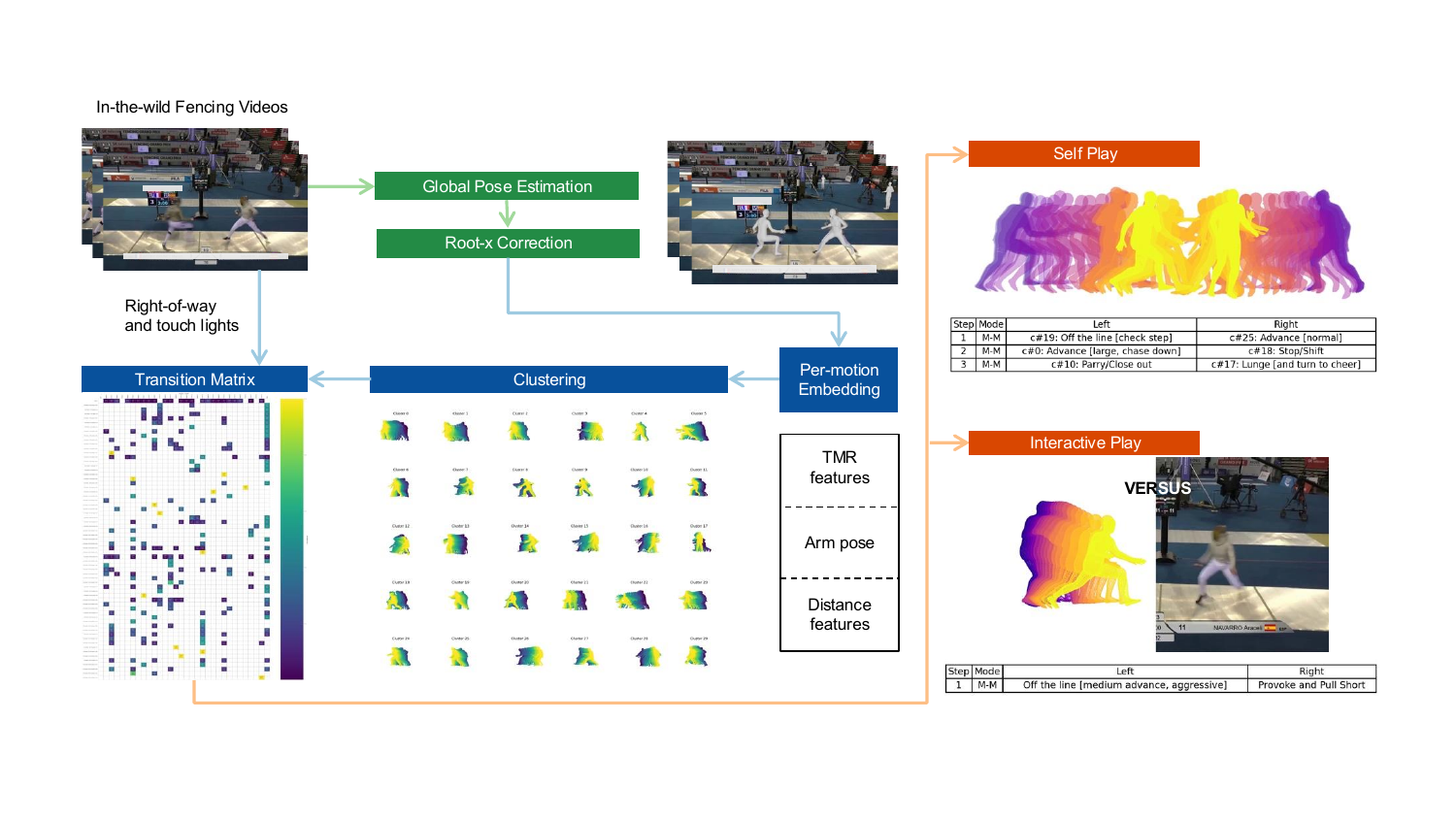}
  \caption{Overall architecture of VirtualFencer. We start with a large, unstructured dataset of in-the-wild fencing videos. Our goal is to extract a reusable representation of strategy from these videos. }
  \label{fig:system_diagram}
\end{figure*}

Our goal is to learn realistic and reusable fencing strategies from unlabeled, in-the-wild video footage. We define strategy as a sequence of actions and the patterns by which fencers transition between them. To achieve this, our system is built around three steps: (a) identifying actions in fencing videos, (b) automatically discovering recurring transition patterns, and (c) enabling the reuse of these patterns to synthesize new touches. At a high level, our pipeline begins by applying off-the-shelf pose estimation methods \cite{wham:cvpr:2024}, which we modify to leverage fencing-specific domain knowledge, to extract a large library of coarse motion sequences from videos of two athletes fencing. In service of (b), we then use unsupervised clustering to group these motion sequences into groups of similar actions. These clusters are reusable: we fit a statistical model over them to learn action probabilities conditioned on contextual variables such as the opponent’s motion and right-of-way. This model captures how fencers typically respond in different situations, enabling us to generate novel touches by sampling plausible sequences of actions. We illustrate this system in Fig.~\ref{fig:system_diagram}.



\section{Video Annotation} 
\subsection{Data Source}
We design a pipeline to collect data from online videos, including world cups, grand prixes, zonal championships, world championships, or any other source of fencing video that adheres to similar quality standards. For this work, we focus specifically on the Senior international circuit hosted by the International Fencing Federation (FIE). We collect $40$ direct elimination bouts, which add up to around 1.5 hours of fencing actions.



For each video, we split it into the individual touches, and extract the 3D motion from both fencers, alongside contextual data such as who has priority and who scored the touch.




\subsection{Tracking and Pose Estimation}
To obtain fencing motions, we use WHAM \cite{wham:cvpr:2024} to reconstruct 3D human poses (SMPL format) in global coordinates from videos. 
We find that using WHAM off-the-shelf provides good-quality local poses but fails at estimating global translations with the accuracy required for studying fencing. 

To correct the global x-axis translations from the monocular video, we estimate a camera homography projection aligned to the canonical fencing piste (14 meters long, 2 meters wide) to project both fencers from pixel coordinates to global coordinates. This is possible because most fencing competitions use a panning camera with fixed position and zoom-levels. 

To obtain the camera homography, we use SAM2 to track the five lines on the fencing piste lines -- left warning, left en garde, middle, right en garde, and right warning -- which are located at 2, 5, 7, 9, and 12 meters from the left end of the piste. Typically at least two lines are within the camera's view, so we are able to compute the homography. 

To get the pixel coordinates of these lines, we use SAM2 to track the fencing piste and the five lines on the piste. For any video, we provide SAM2 five prompt points of a frame, then we propagate these five instance maskings throughout the video. Because fencing is fast, some frames can be so blurry that SAM2 only tracks a partial line. To solve this, we isolate out the piste borders by reapplying the union of the five prompt points to segment out the fencing piste. We then run an LSD-based line detector on this piste and prune by selecting the pair that best aligns with the top and bottom contours of the piste mask out of all near-horizontal lines. With the pixel coordinates of these lines, we are able to obtain a homography-based projection from camera space to the world space. 

To obtain a fencer's pixel coordinates, we extract their representative 2D position on the piste using the following procedure. First, we run YOLO detection on any frame to obtain prompt points, which are then passed into the SAM2 tracking  to propagate through the video. Then, for every frame, we compute the median x-coordinate of their segmentation mask and cast a vertical ray through the median x-coordinate to intersect it with the two piste borders, and the midpoint between two intersections is used as the fencer’s on-strip pixel coordinate.


\subsection{Metadata: Annotate Priority and Identify Scoring Lights}
\label{sec:priority}
\paragraph{Priority} To support downstream strategy modeling, we annotate each fencing touch with an estimated priority mode every 20 frames. When the touch begins, no one has priority (mode M-M). For each of the following 20 frames, we compare the two fencers' displacements to infer initiative based on which fencer closed the relative distance more aggressively. Let $\Delta x_L$ and $\Delta x_R$ be the displacements of the left and right fencer in the local forward direction. We compute the difference in displacements $\Delta = \Delta x_L - \Delta x_R$. If $\Delta > \delta$, meaning the left fencer noticeably moved forward more than the right, priority is assigned to the left fencer in the next timestep; if $\Delta < -\delta$, priority is assigned to the right fencer at the next timestep. Otherwise, the previous priority mode is retained. We use $\delta = 0.3$ in all experiments. 

\paragraph{Scoring Lights} For every clip, we annotate the timestep when either of the fencers scores a touch, which is indicated by a red (for left) or green (for right) light at the bottom of the online video data.

\section{Strategy Extraction} 
\subsection{Motion Embedding}
While there is a finite number of actions a fencer might choose to use, these actions require expert knowledge to identify. Moreover, the need for quick action and reaction means that even a short clip might have several actions in it, making the task of manually annotating fencing videos very burdensome. Based on our measurements, we find that the typical duration of a fencing action is around $20$ frames. Building on this insight, we divide each fencing clip into consecutive 20-frame subclips. For each subclip, we compute a motion embedding by concatenating three types of features:

\begin{itemize}
    \item Text-to-motion retrieval (TMR \cite{petrovich23tmr}) embedding
    \item the axis-angle representation of the dominant arm's elbow and wrist joints
    \item global distance features
\end{itemize}

We find that the features from TMR are particularly effective for differentiating both coarse actions (lunges and steps, for example) and variations of execution of the same action (aggressive steps and patient steps, or prep steps and hop steps). They are also useful for differentiating based on modes of fencing footwork, which has been a major focus of prior research \cite{MALAWSKI20181, MALAWSKI2019198, Zhu_2022_CVPR}.

We explicitly include the pose of the fencer's dominant arm joints (elbow and wrist) because dominant arm movements (like hit and parries) are central to fencing exchanges.

The use of axis-angle representation emphasizes rotational changes, enabling us to distinguish between similar footwork executed with different arm postures (stepping forward with the arm raised versus held back, for example).

The global distance feature includes:

\begin{itemize}
    \item net displacement (the difference between the fencer's current position and start position)
    \item the start and end zone along the strip, discretized into five regions (left warning zone, left en garde zone, middle zone, right en garde zone, and right warning zone)
    \item the maximum forward and backward displacement over the 20-frame window
    \item the ratio of peak to median speed
\end{itemize}

These elements capture tempo shifts and positional dynamics in fencing movement, such as in-and-out motions, or sudden stopping within the box (which are common tactics used to disrupt the opponent's rhythm). 

\subsection{Low-level Skill Discovery}
\label{sec:clustering}
We adopt a two-stage unsupervised clustering process to identify meaningful low-level skills from motion embeddings. 

Due to the scale of our dataset, constructing a global clustering model is computationally expensive.
Instead, we manually select 10 bouts (2210 samples) to serve as representative samples. We ensure that these span a wide range of fencing styles by including both men’s and women’s events, as well as fencers from different countries to account for regional differences such as Asian versus European styles.

For the first stage, we run the k-means clustering algorithm on this subset, manually inspect each cluster, and identify the ones that do not contain any action (for example, standing still before a touch starts). We discard the samples that fell into these clusters and cluster the remaining ones again. After validation, the clusters found in this second clustering stage are used as our discrete fencing actions.




We evaluate the resulting cluster quality in Section~\ref{sec:cluster-eval}. At inference time we can retrieve the action corresponding to a given sample by finding the cluster it belongs to. Alternatively, given an action, we can retrieve a clip from the corresponding cluster.

\subsection{High-level Strategy Learning}
\label{sec:core_hyp}
We model fencing strategy as a variable-length walk over low-level action clusters that terminates upon the bout-ending conditions described in Sec.~\ref{sec:finish_cond}, reflecting a fencer's evolving strategic decision-making over time.

Our core hypothesis is that, given the amount of structure that we describe next, we can accurately learn fencing strategy directly from data.

We fit three separate conditional transition matrices under three priority conditions: \textbf{M-M} (at the start both fencers are in the \textbf{m}iddle, and neither has priority), \textbf{P-NP} (the fencer has priority), and \textbf{NP-P} (the opponent has priority). Modeling these scenarios separately is essential because a fencer's tactical behavior varies significantly under different priority conditions: when neither fencer has priority, they will fight to get it; when holding priority, a fencer will act more conservatively to avoid making mistakes; when aiming to take priority from the opponent, a fencer will freely provoke the opponent in order to cause a mistake.


We model high-level fencing strategy as a sequence of low-level actions, each drawn from the discrete set of 30 motion clusters $\{C_0, C_1, \ldots, C_{29}\}$ described in Section~\ref{sec:clustering}. Let $U$ and $V$ denote the two fencers in a bout. At each timestep $t$, fencer $U$ executes action $u_t$ and opponent $V$ executes $v_t$. Let $d_t$ denotes their relative distance on the strip.

We define a priority-aware \textbf{strategy model} as a probability function,
$$
P(u_t \mid u_{t-1}, v_{t-1}, d_t),
$$
that captures the probability that fencer $U$ selects action $u_t$ given their own previous action $u_{t-1}$, the opponent’s previous action $v_{t-1}$, and the current relative distance $d_t$. We hypothesize that, due to the fast and reactive nature of fencing, fencers primarily attend to their opponent’s most recent action while maintaining consistency with their own prior motion, rather than to the full trajectory that led to the current moment. As such, a single step of history is sufficient to capture realistic temporal dependencies.

We model this distribution as
\begin{equation}
\begin{aligned}
P(u_t \mid u_{t-1}, v_{t-1}, d_t) \propto\ 
& P_{\text{raw}}(u_t \mid u_{t-1}, v_{t-1}) \\
& \cdot \exp\left( -\frac{1}{2} \left( \frac{d_t - \bar{d}(u_{t-1}, v_{t-1})}{\sigma} \right)^2 \right),
\end{aligned}
\end{equation}
where the first term is the probability distribution over actions given the prior action of the fencer and the opponent (which we will refer to as raw transition distribution) and the exponential term is a distance-aware weighting term that adjusts the probabilities of the actions based on the distance between the two fencers.

We use the fencing trajectories in our dataset to estimate the raw transition distribution. Given sequences of the form

\[
\{(u_0, v_0, d_0), (u_1, v_1, d_1), \ldots, (u_T, v_T, d_T)\},
\]
we compute the raw transition distribution $P_{\text{raw}}(u_t \mid u_{t-1}, v_{t-1})$ from empirical frequency counts (that is, how often was action $u_t$ chosen given $u_{t-1}$ and $v_{t-1}$). We also compute the average distance $\bar{d}(u_{t-1}, v_{t-1})$ by averaging the distances between fencers at the moment when each transition is observed.

Distance awareness is introduced at inference time by weighting the raw transition probabilities with a Gaussian function of the current distance $d_t$. For a given transition context $(u_{t-1}, v_{t-1})$, we retrieve the corresponding raw transition distribution and average distance $\bar{d}$. This weighting function has a tunable parameter $\sigma$ which we can modify to calibrate the sensitivity of our model to the distance between fencers (we set $\sigma = 0.5$ in all experiments). This continuous distance-based bias allows the model to capture position-dependent strategy variation (for example, fencers are more likely to initiate a lunge when at the optimal attack range, and more likely to take preparatory steps when still too far to engage). 

This process is repeated independently for each of the three priority modes, resulting in three distinct raw transition distributions that factor priority awareness into our fencing strategy model.

\subsection{Simulating a Fencing Touch}

We simulate a fencing touch as a turn-based interactive sequence between two fencers, each reacting to the other under evolving priority constraints.
Our simulation system operates as a state machine. At each timestep, the system infers the current priority mode based on the previous pair of motion sequences, samples an action for each fencer from the strategy model associated with that mode, retrieves a motion sequence from the selected cluster, updates the simulation state according to the outcome of the interaction, and evaluates whether a termination condition has been met.

\subsubsection{Priority Heuristic}
Each touch starts with an M-M priority mode. At every timestep, the priority mode is updated by a transition function that reflects fencing rules, evaluating the outcome of the two executed motion sequences. Specifically, it follows three cues:

\paragraph{Metadata.} For each sample in our dataset we use the information from the scoring light to see if it led to a point or not. As such, we have this information available at inference time. If only one fencer’s executed motion sequence corresponds to a scoring light in the original bout, but the bout does not fulfill the termination conditions, priority is awarded to the opponent in the next timestep. This models the rule that a miss forfeits right-of-way.

\paragraph{Finishing actions.} In the absence of lights, we manually select a set of clusters whose actions are usually associated with offensive intent to be considered as finishing actions. If only one fencer performs such an action and the finishing conditions are not met, we assign priority to the opponent in the next timestep.

\paragraph{Displacement.} If all of the above fail, we fall back to the Section~\ref{sec:priority} to compare.

When priority modes switch, we randomly sample a starting action rather than condition on prior transitions. This reflects how fencers “reset” their internal planning when priority switches, as the way priority was gained or lost has little bearing on how they act under the new mode.

\subsubsection{Finishing Condition}
\label{sec:finish_cond}
In our implementation, simulation stop conditions are inspired by real life fencing rules that dictate when a point should terminate. The simulation terminates if one the following conditions are met:

\begin{itemize}
    \item \textbf{Out of bounds:} either fencer steps off the strip.
    \item \textbf{Crash:} the distance between the global root positions of the two fencers falls below a threshold $\tau=1.5$ meters, suggesting a collision
    \item \textbf{Touch registered:} the fencers are within 2 meters of each other, and a scoring light is triggered in either of the sampled motions, indicating a valid touch and end to the point
    \item \textbf{Terminal action:} the fencers are at distance (< 2 meters) and a fencer executes a pre-defined finish cluster.
\end{itemize}

These stopping conditions reflect plausible bout termination criteria consistent with real life fencing.

\section{Results}
\label{sec:results}

\subsection{Skill clustering}
\label{sec:cluster-eval}
We evaluate the quality of our clustering quantitatively using an estimate of clustering accuracy. After our clusters are done, we inspect them and assign them labels corresponding to the fencing actions seen in the cluster samples. We label each cluster with a standard general category of fencing motion (Advance, Retreat, Lunge, and Off the line) and a secondary label providing more detail within that general category. After each cluster is labeled, we classify a test set of samples. Given the cluster labels for the test set, we randomly select 10 test samples per cluster, and manually check if the action in those samples matches the cluster label. Average clustering accuracy is 85.67\%. We report cluster action names and full cluster-level accuracies in Table \ref{tab:cluster_info}.

Our clustering approach correctly groups together samples with the most common actions seen in saber fencing, such as lunges and parries. We experiment with the number of clusters and find that using 30 maintains this grouping, while unlocking finer distinctions between fencing actions. For example, two clusters (2 - Off the line [prep steps] and 19 - Off the line [check step]) both have clips with small preparation steps, but in one the motion starts with the front foot, while in the other it starts with the back foot. At the same time, our clustering approach correctly classifies variations of the same action across diverse athletes within a reasonable range of body builds (one limitation we encounter is that a small step for the tallest athlete might be considered a large step for the shortest one). Finally, there is little redundancy in the clusters our method yields, with each cluster adding additional resolution into the fencing motion space.

\subsection{Emergent Strategy}

\begin{figure*}[!t]
  \centering
  \includegraphics[width=\textwidth]{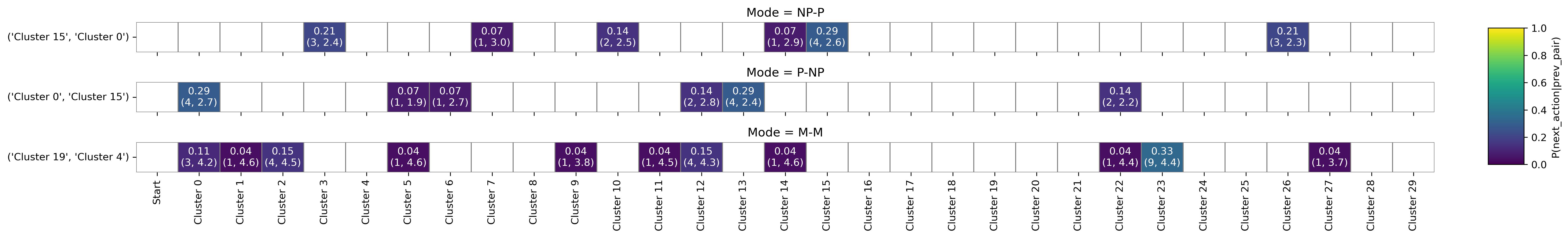}
  \caption{Selected rows of the transition matrices, from top to bottom:\\
  \textbf{NP-P condition} -- The left fencer has just retreated (C.15), and the right fencer has just advanced (C.0). Even though the most typical and rudimentary action from this state is another simple retreat, the top row shows that the system chooses from a variety of more nuanced and equally appropriate ways to move on defense, ranging applying short-distance pressure (C.15) to provoking and pulling away (C.7).\\
  \textbf{P-NP condition} -- The left fencer has just advanced (C.0), and the right fencer has just retreated (C.15). Even though the most typical action from this state is another advance, the middle row shows that the system chooses from a variety of more nuanced ways to complete the attack, such as attacking aggressively (C.22) versus passively (C.13).\\
  \textbf{M-M condition} -- The left and right fencer have both just taken preparation steps forward (C.19,4) and neither side has priority. The bottom row shows that when there is not necessarily a clear next optimal action, the system offers a broad variety of possible actions, from highly offensive (C.1,5) to highly defensive (C.12,14).
  }
  \label{fig:transition_matrix_select}
\end{figure*}

Figure~\ref{fig:transition_matrix_select} shows a slice of the learned strategy for three different contexts (pairs of previous actions) and priority conditions. We refer to the image caption for a detailed description of the strategy. From a high-level perspective, our model learns to consider several possible options that are reasonable given the context and the priority (such as advancing again when having priority, or attacking in different ways).


\subsection{User Study}
To assess whether our system's understanding of strategy reflects real-world fencing tactics, we conducted a user study with current and former NCAA Division 1 saber fencers--the highest level of collegiate competition in the United States--many of whom have competed on the senior international circuit. We run three experiments to evaluate three different applications of our system. In the first two, participants are shown several sets of videos, each set consisting of three conditions (action predictions/touches): one generated by our system, one generated at random, and one corresponding to the ground truth. Participants did not know which condition each video belonged to and were asked to rank the options from most to least strategy-abiding. For the third experiment, we had users fence directly against our system.  



\subsubsection{Fence against an online trajectory (next action prediction)}
\begin{figure*}[!htbp]
  \centering
  \begin{minipage}[t]{0.3\textwidth}
    \centering
    \includegraphics[width=0.3\linewidth]{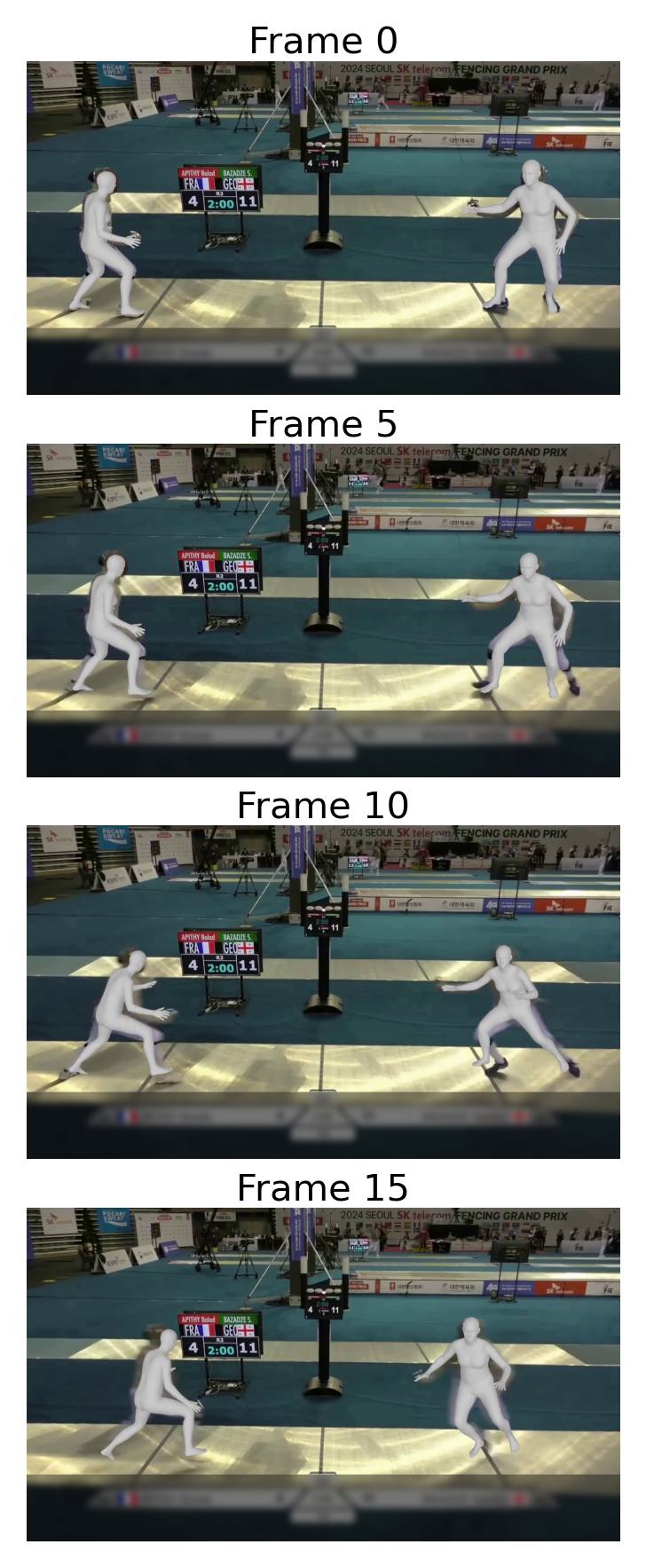}
    \includegraphics[width=0.45\linewidth]{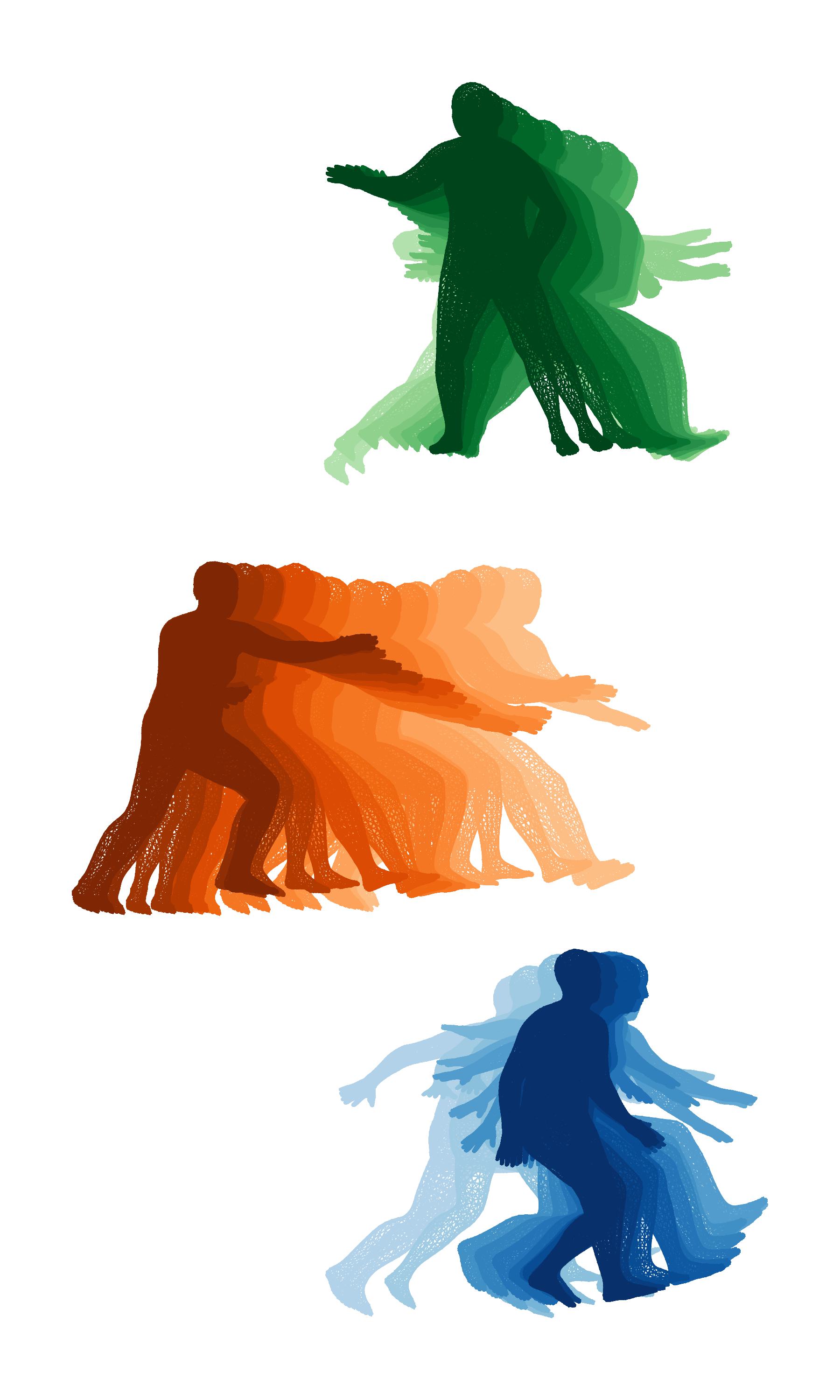}
    \caption*{A sample under mode M-M.}
  \end{minipage}\hfill
  \begin{minipage}[t]{0.3\textwidth}
    \centering
    \includegraphics[width=0.3\linewidth]{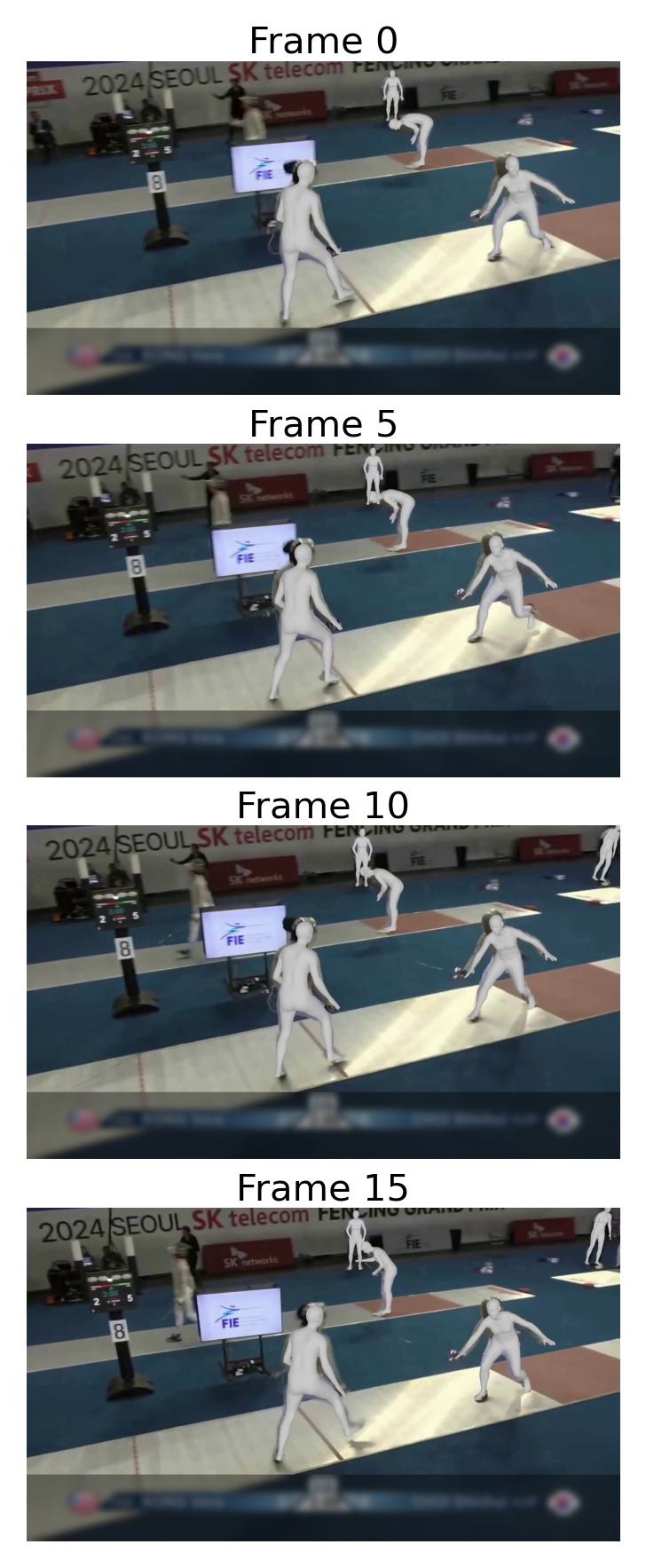}
    \includegraphics[width=0.45\linewidth]{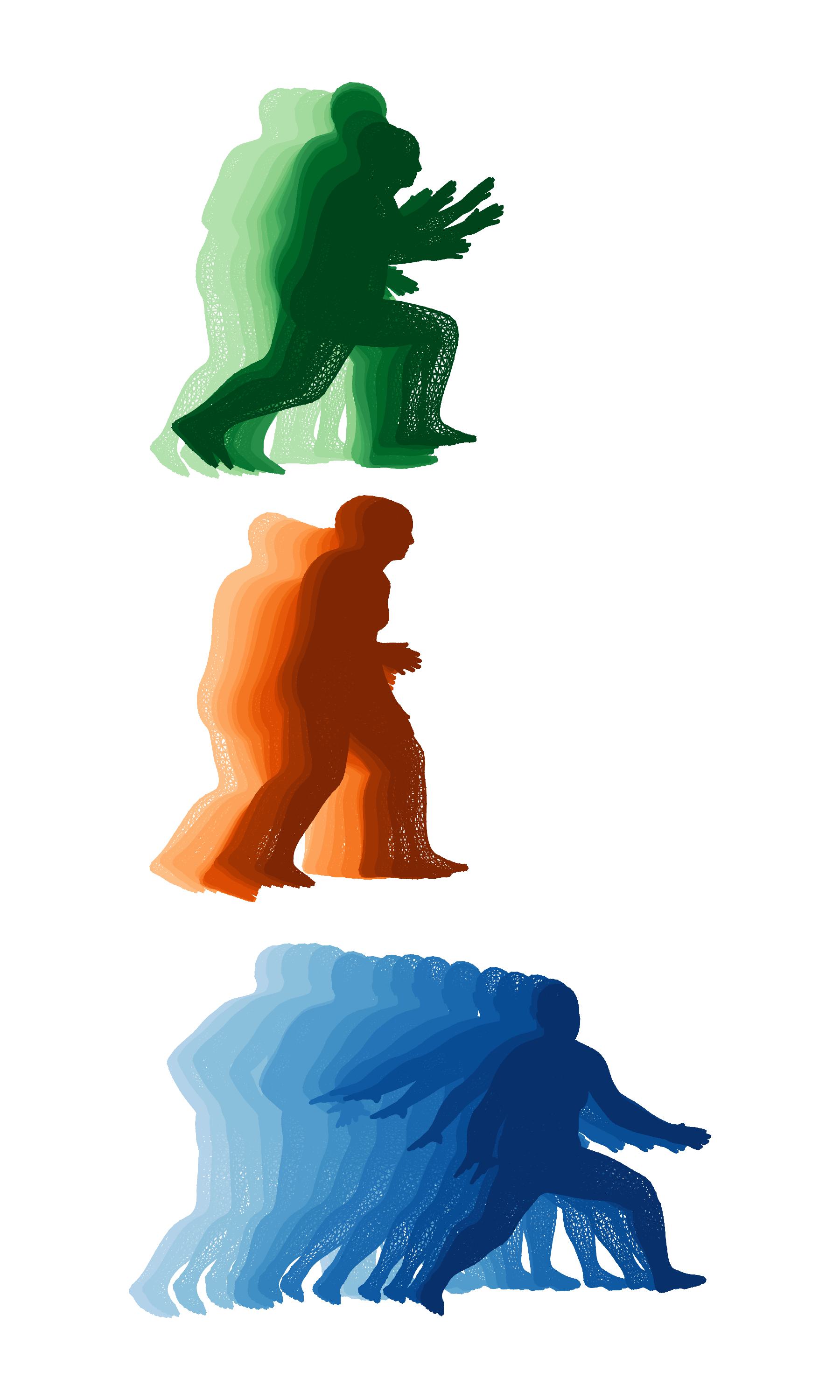}
    \caption*{A sample under mode P-NP.}
  \end{minipage}\hfill
  \begin{minipage}[t]{0.3\textwidth}
    \centering
    \includegraphics[width=0.3\linewidth]{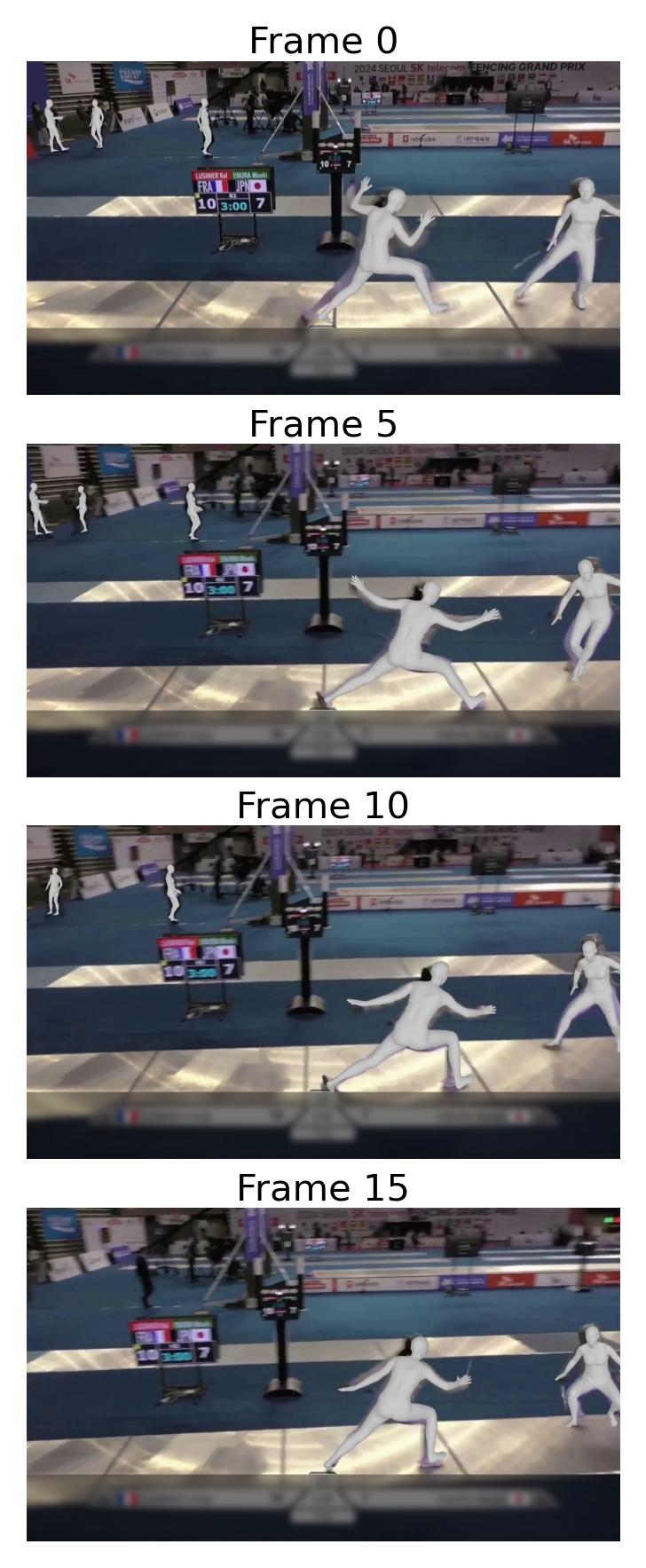}
    \includegraphics[width=0.45\linewidth]{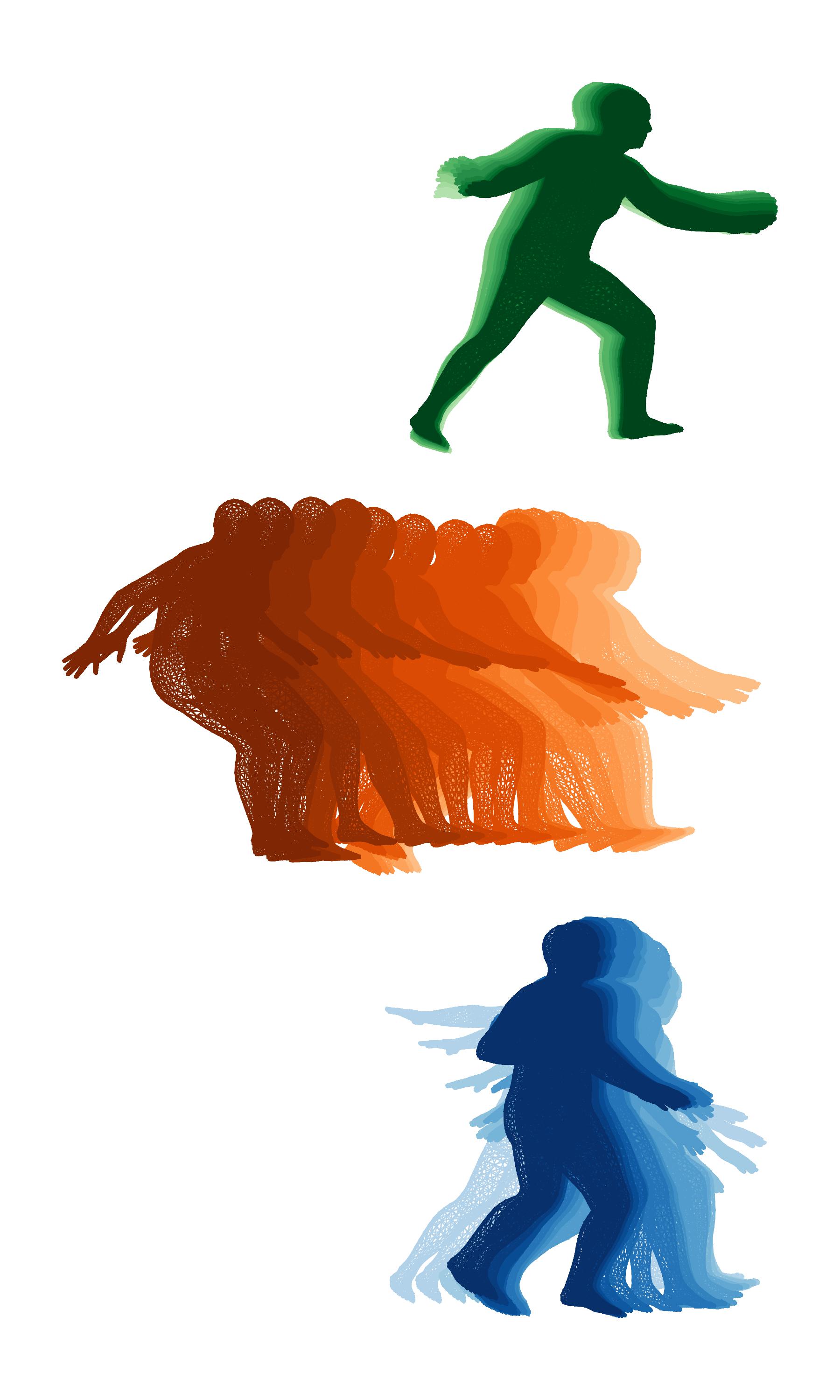}
    \caption*{A sample under mode NP-P.}
  \end{minipage}

  \caption{First User Study. Given the context of the in-progress touch, users are asked to rank the next action choices from ground truth (green), random (orange), and our VirtualFencer (blue) as if they were the left-hand-side fencer. Shading from light-to-dark indicates time progression.}
  \label{fig:user_study_2}
\end{figure*}

First, we asked 23 participants to judge our system's choice of the next action in response to a real fencer's actions. Given a video clip of part of a touch, participants were shown the next action as picked by each of the three conditions (see Figure~\ref{fig:user_study_2}). For picking the VirtualFencer action, we queried it given the context of the touch. The participants were asked to choose among three options (the ground truth, the action chosen by VirtualFencer, and a random action) as the response for the left-hand-side fencer. Each participant viewed 25 sets of 3 videos each for a total of 575 data points. The results are shown in Table~\ref{tab:user_study_table}. Users showed a clear preference towards the ground truth motion, followed by the action picked by VirtualFencer, and finally the random baseline.

This study provides us with some measure of how good of a next action predictor VirtualFencer is. However, a single action is not enough to showcase the learned strategy, and fencers from online video cannot react to our system, which motivates the next user study.

\subsubsection{Fence against itself (self-play)}
\begin{figure*}[!htbp]
  \centering
  \begin{minipage}[t]{0.32\textwidth}
    \centering
    \includegraphics[width=\linewidth]{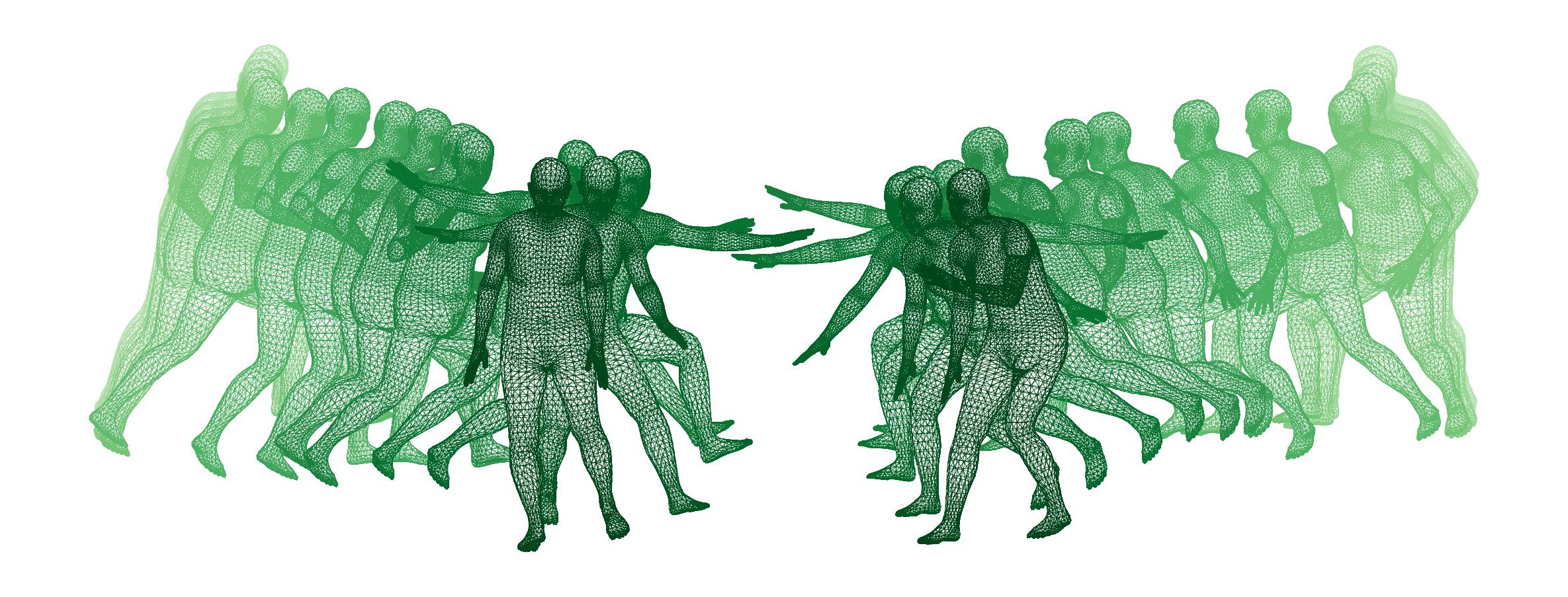}
    \caption*{Ground Truth}
  \end{minipage}
  \hfill
  \begin{minipage}[t]{0.32\textwidth}
    \centering
    \includegraphics[width=\linewidth]{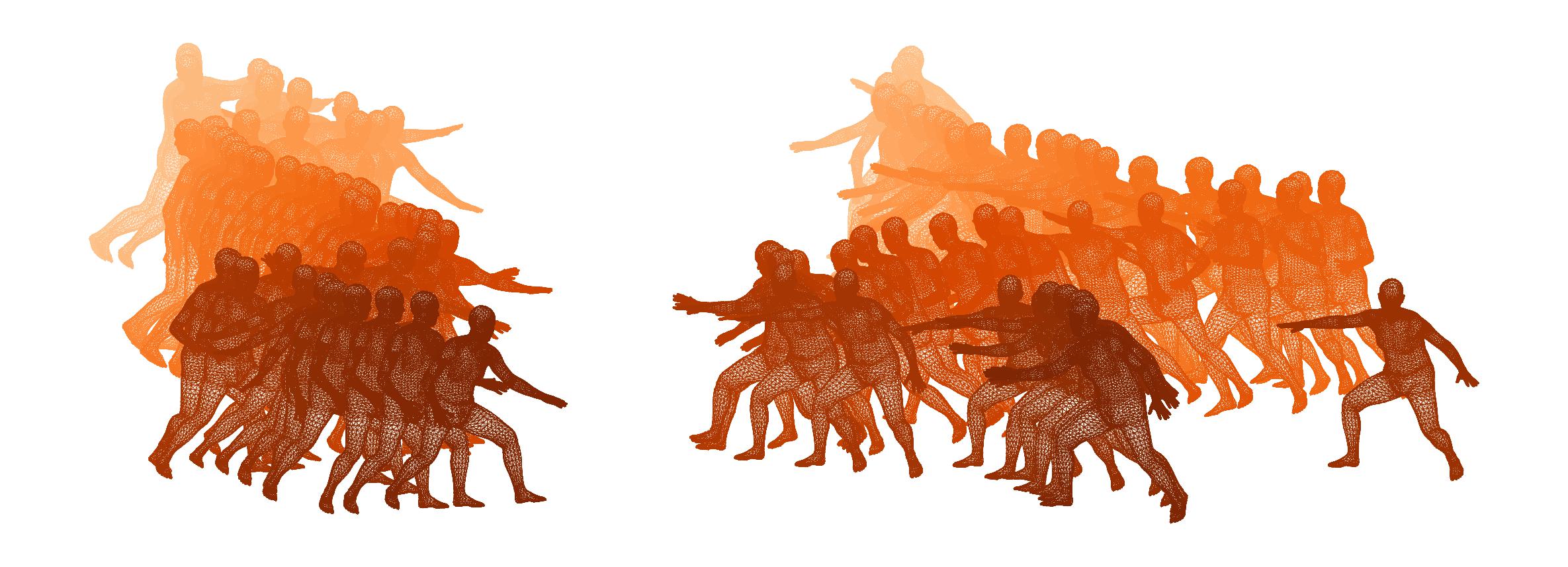}
    \caption*{Random}
  \end{minipage}
  \hfill
  \begin{minipage}[t]{0.32\textwidth}
    \centering
    \includegraphics[width=\linewidth]{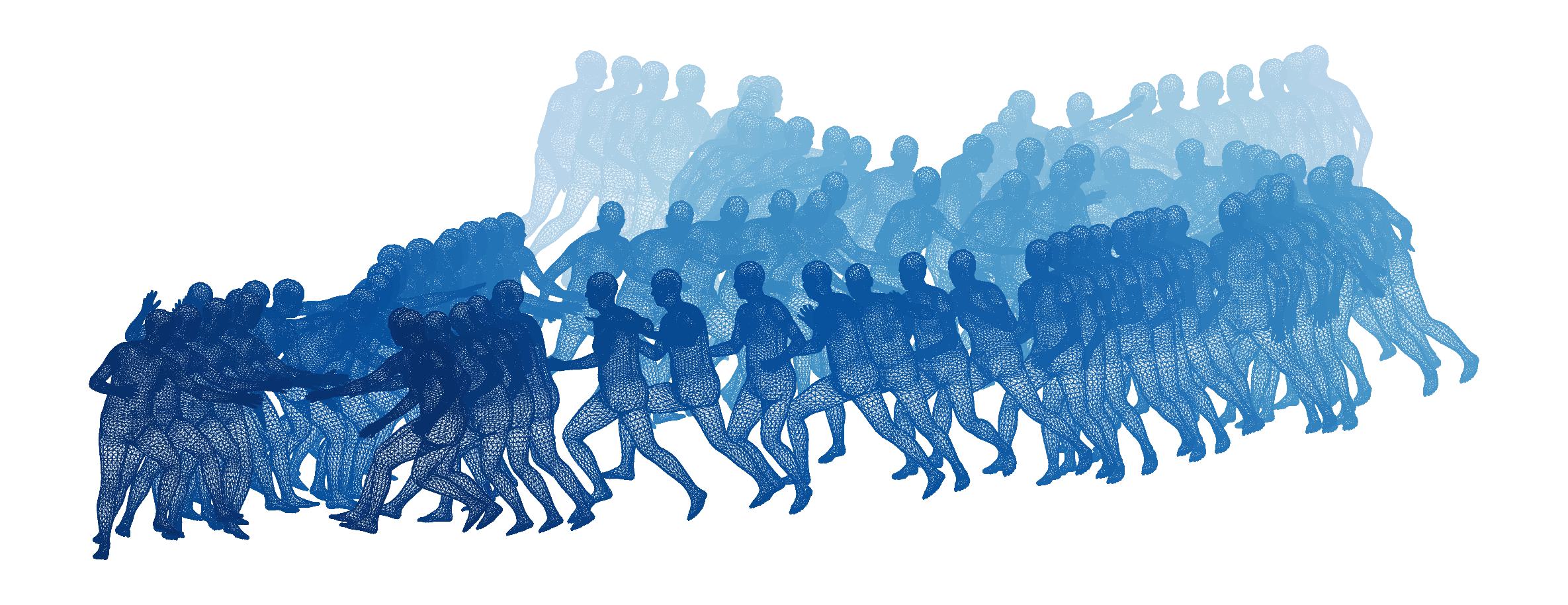}
    \caption*{Ours}
  \end{minipage}
  \caption{Second User Study. Users were shown temporal motion rollouts for three conditions -- from left to right: sampling cluster trajectories using ground truth, random, and our model -- and asked which touch they preferred. Each figure visualizes a full touch, where top-to-bottom and light-to-dark indicate time progression.}
  \label{fig:user_study_1}
\end{figure*}

To evaluate the validity of our system's sequential decision-making, we asked 20 participants to watch and evaluate 25 sets of videos (a total of 500 data points) generated through self-play.  Figure~\ref{fig:user_study_1} visualizes representative bouts. Table \ref{tab:user_study_table} shows the ranking results. This time, users split their preferences evenly between touches produced by VirtualFencer's self-play and ground truth touches, while clearly preferring both methods to the random baseline. This shows that while our system might not be the preferred next action predictor (as evidenced by the first user study), when chaining several actions together it generates a touch that is good enough to attract the preference of the users.


\begin{table}[t]
\centering
\caption{Distribution of votes per rank by player for both user studies}
\begin{tabular}{lrrrlll}
 & \multicolumn{3}{c}{First user study}& \multicolumn{3}{c}{Second user study} \\
\cmidrule(lr){2-4}\cmidrule(lr){5-7}
Player & 1st & 2nd & 3rd  & 1st & 2nd &3rd   \\
\cmidrule(lr){1-1}\cmidrule(lr){2-4}\cmidrule(lr){5-7}
VirtualFencer     & 154 & 219 & 202   & 218 & 172 &110 \\
Ground Truth   & 300 & 173 &  102   & 177 & 188 &135 \\
Random   &  121 & 183 & 271   & 105 & 140 &255 \\
\cmidrule(lr){1-1}\cmidrule(lr){2-4}\cmidrule(lr){5-7}
\end{tabular}
\label{tab:user_study_table}
\end{table}
%




\subsubsection{Fence against professional fencers}

Finally, we conducted an expert evaluation where four professional fencers, each with over 10 years of competitive experience, engaged in real-time, interactive exchanges with VirtualFencer. With their own distinct style, each of the participants probed different tactical corners of our system. Each session involved fencing to five touches. Participants were instructed to think strategically, taking factors such as priority, distance, and timing into account. At each timestep, both the participant and our system were prompted to select one of the 30 actions available. 

Through these sessions, we gained qualitative insights into VirtualFencer's strengths and limitations. The experts noted that while VirtualFencer demonstrated understanding of mature tactics like the "pacing of a long attack", it still made beginner errors like lunging when it did not have priority. Additional expert feedback is shared in Table \ref{tab:vf-feedback}.

On a scale from 1 to 5, the experts gave the system an average of 4 for understanding priority, 4.25 for action diversity, and 3.75 for maintaining an appropriate distance. The lower score for distance can be attributed to the system's occasionally "unreasonable" decisions (such as lunging without priority) despite accurately detecting distance cues. Experts placed VirtualFencer's understanding of strategy somewhere between the Y14 to Junior (Y20) level. These observations both validate our proposed system's capability to learn strategy from in-the-wild video and also highlight directions for future work. 







\section{Conclusion}

We introduce VirtualFencer, a system for extracting fencing motion and strategy from in-the-wild video data. Our system recovers the 3D motion of two athletes fencing each other and uses a two-phase clustering approach to extract low-level fencing skills. Using our low-level skills and our dataset of fencing motion, we extract high-level fencing strategy, which we use to generate novel bouts. We run a user study where we ask professional fencers to rank the bouts generated by our approach relative to real bouts and randomly generated bouts, and find that these trained experts are not able to distinguish between bouts generated by our method and ground truth bouts, thus validating the strategy that we reconstruct and providing evidence to support our core hypothesis from Section~\ref{sec:core_hyp}.

Our method comes with several limitations and exciting related directions for future work. We have noticed that different actions can be grouped into similar groups based on the confounding variable of an athlete's build. For example, a ``small step'' taken from a tall fencer and a ``large step'' taken from a short fencer may be grouped together due to their similar absolute distances traveled, even though they reflect different intentions. Our hypothesis is that this is because our clusters are built from an intentionally mixed video dataset that includes motions from a wide range of fencers and body types. We believe there is interesting work to be done on fine-tuning clustering based on specific player styles or physical attributes. In general, we believe that incorporating retrieval-based metrics to selectively retrieve motion data for building personalized or task-specific clusters would be an exciting direction for future work.
Some participants in our user study reported difficulty following the rendered videos, largely due to discontinuities between consecutive skill clips. Synthesizing higher-quality motion with smoother transitions, or even exploring fully generative approaches that produce continuous motion sequences directly from a sequence of skill IDs, could improve the coherence and readability of the output. This challenge is related to a broader body of work on generating high-quality motion from low-quality inputs like video~\cite{zhang2023vid2player3d}, and suggests that similar techniques could be leveraged to improve motion realism in our setting.
Our focus in this project was to learn and model realistic fencing strategies at the level of individual points. We did not consider how strategies might evolve over longer timescales, such as across multiple points or throughout an entire bout. However, modeling these longer-term dynamics presents an exciting direction for future work. We are particularly interested in exploring how learned strategies evolve when fencing agents are given access to extended context, such as prior points, score differentials, or fatigue, which could reveal richer tactics and behavior.




\bibliographystyle{ACM-Reference-Format}
\bibliography{aa_main}

\clearpage

\begin{figure*}[t]
  \centering
  \setlength{\tabcolsep}{1pt}
  \renewcommand{\arraystretch}{0.0}
  \begin{tabular}{ccccc}
    \includegraphics[width=0.18\textwidth]{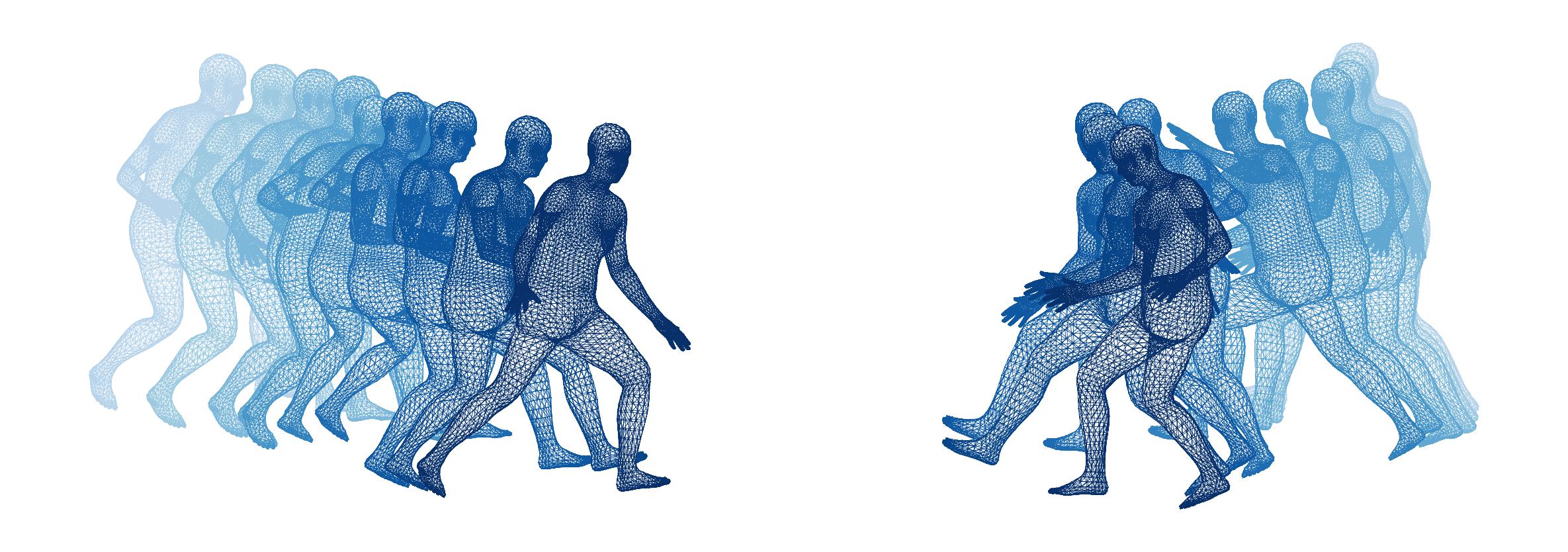} &
    \includegraphics[width=0.18\textwidth]{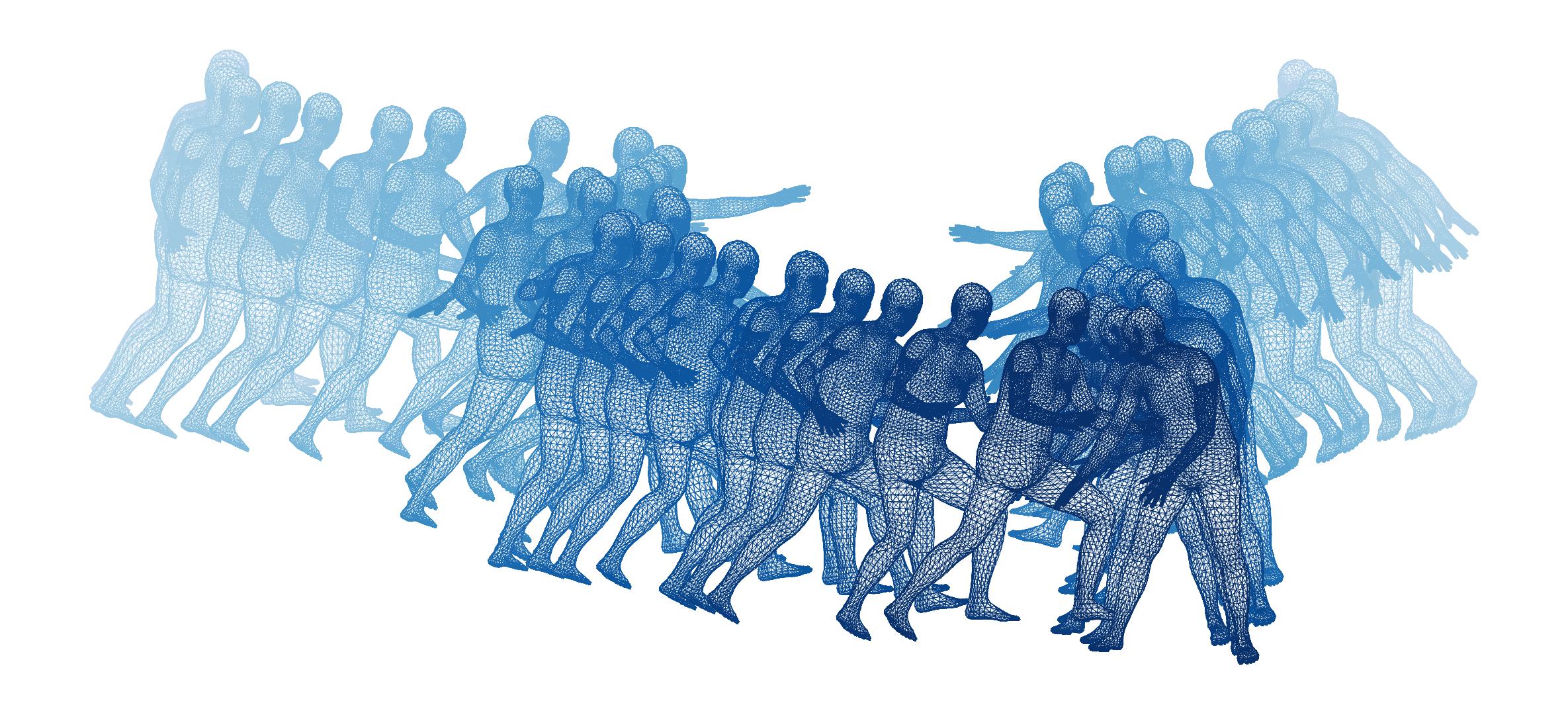} &
    \includegraphics[width=0.18\textwidth]{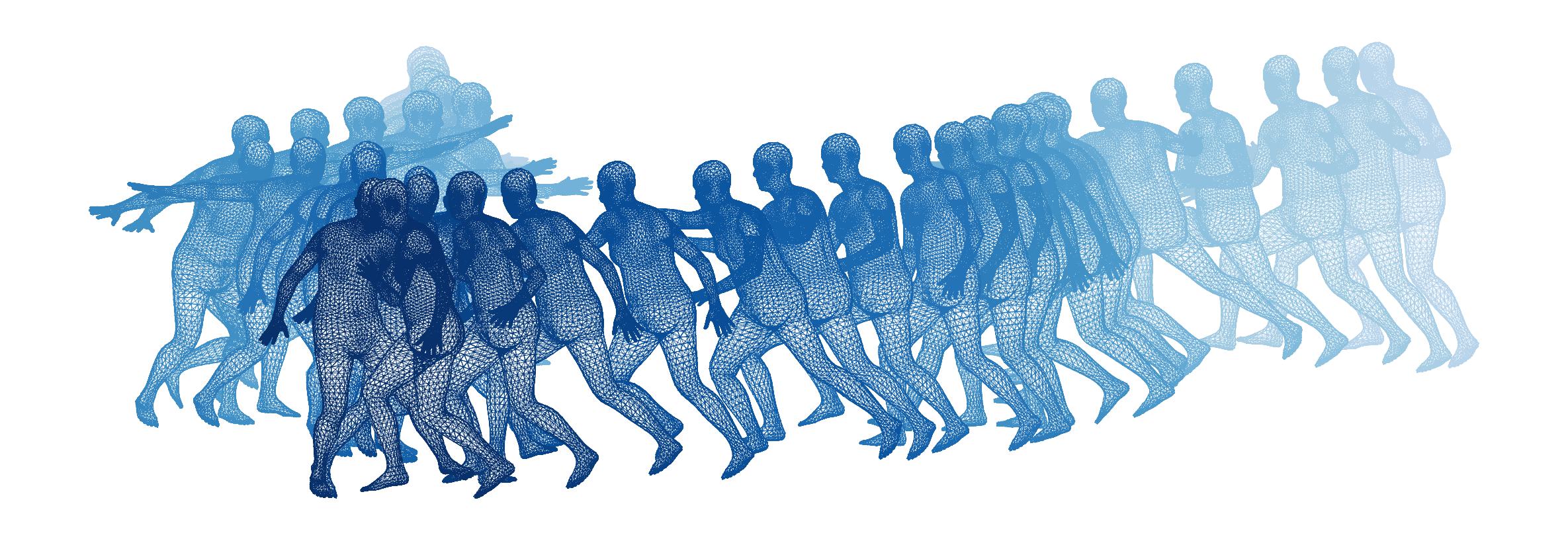} &
    \includegraphics[width=0.18\textwidth]{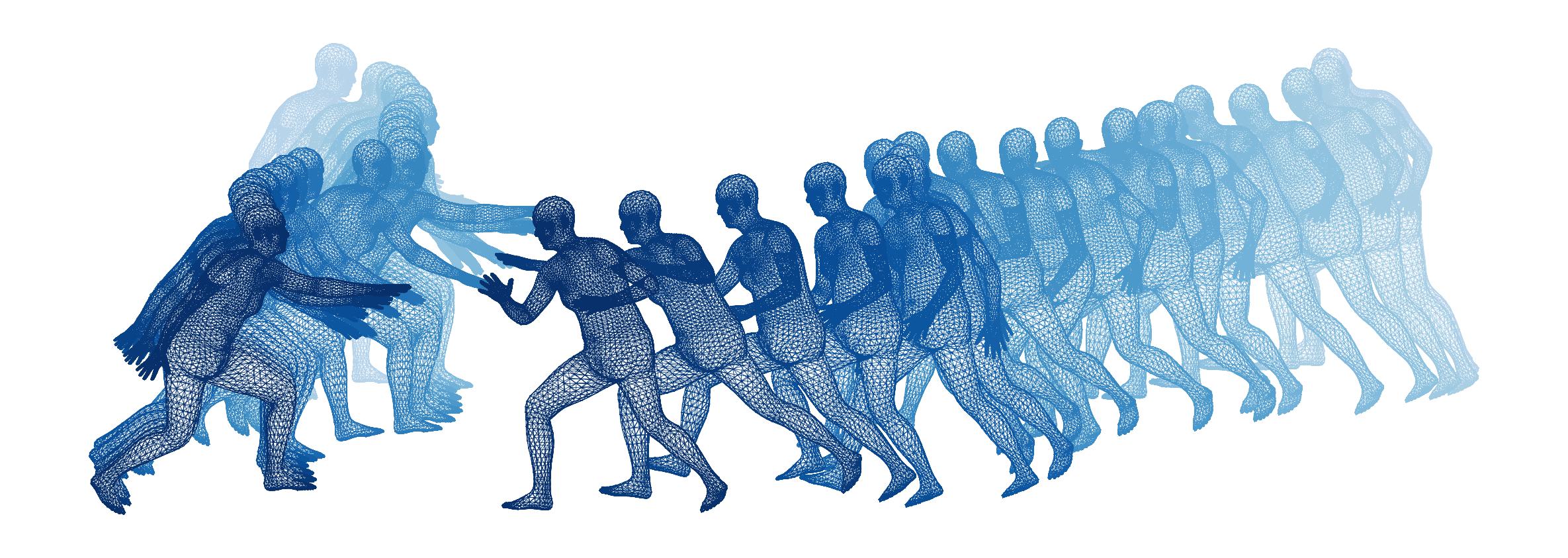} &
    \includegraphics[width=0.18\textwidth]{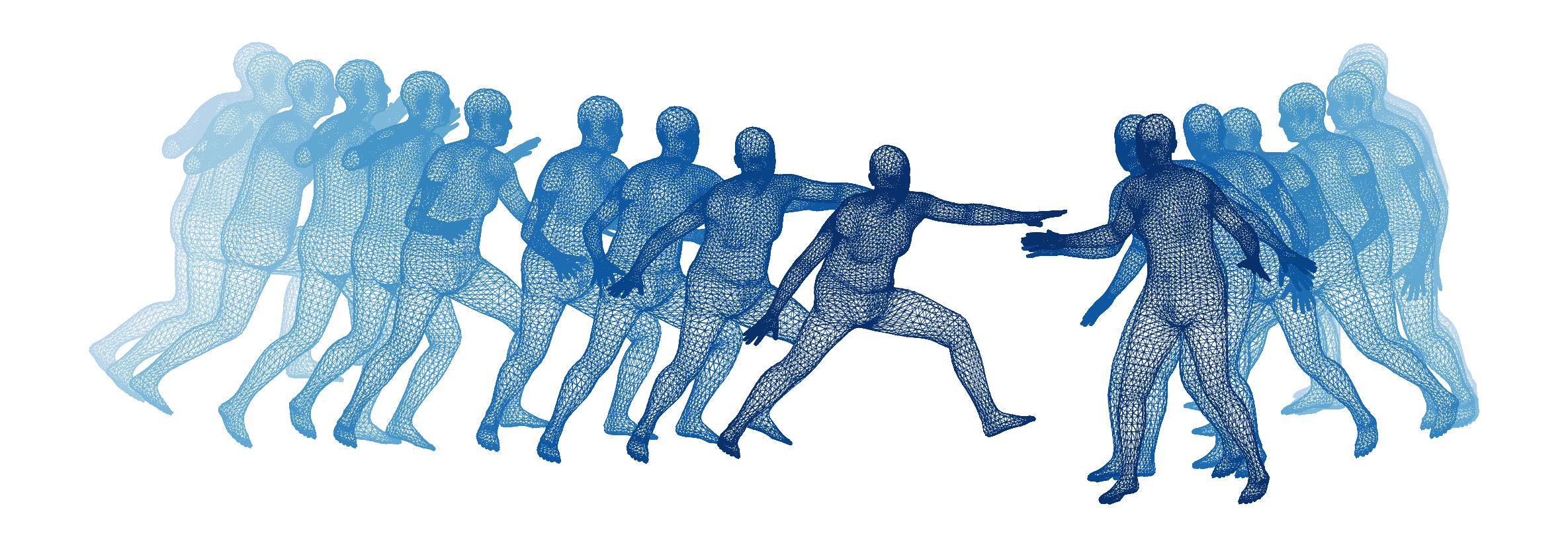} \\
    
    \includegraphics[width=0.18\textwidth]{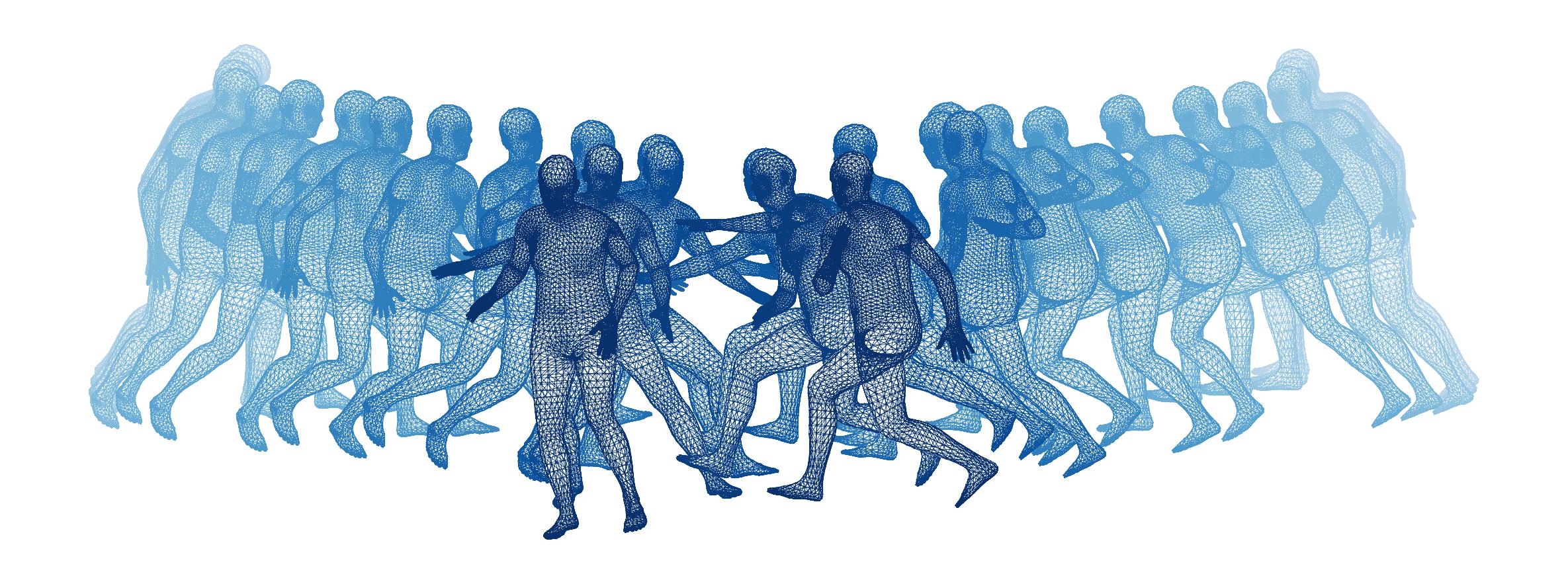} &
    \includegraphics[width=0.18\textwidth]{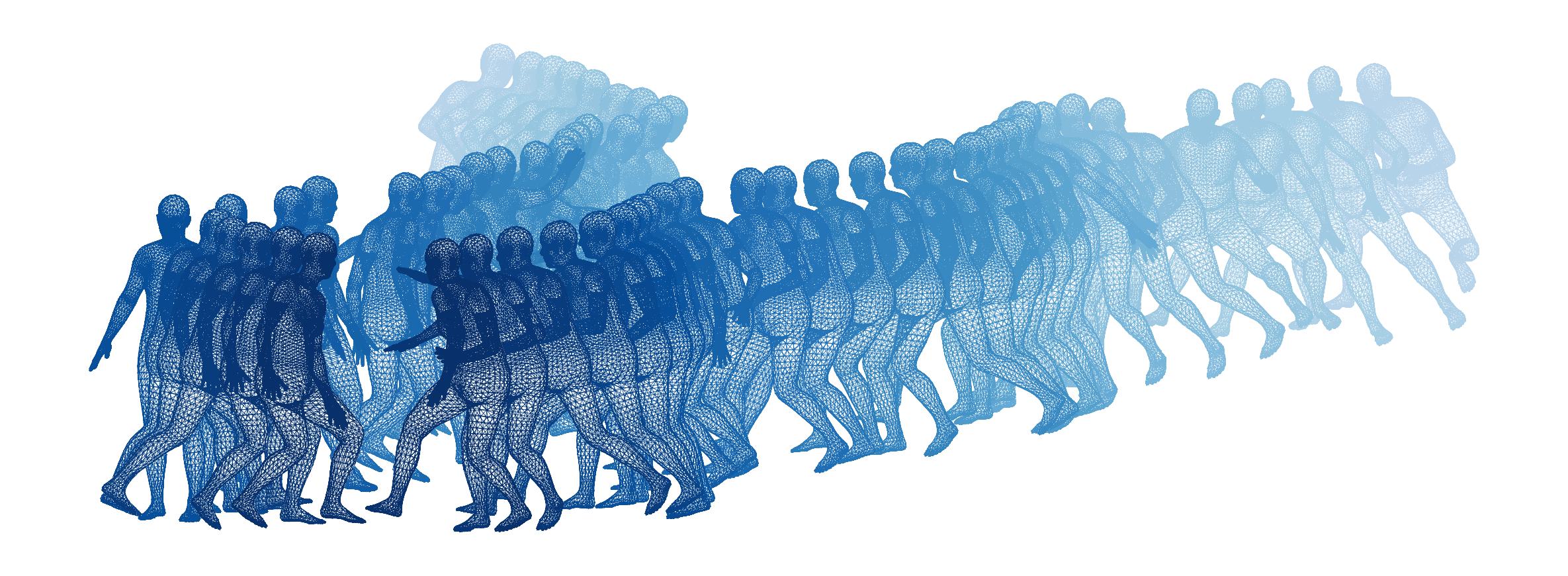} &
    \includegraphics[width=0.18\textwidth]{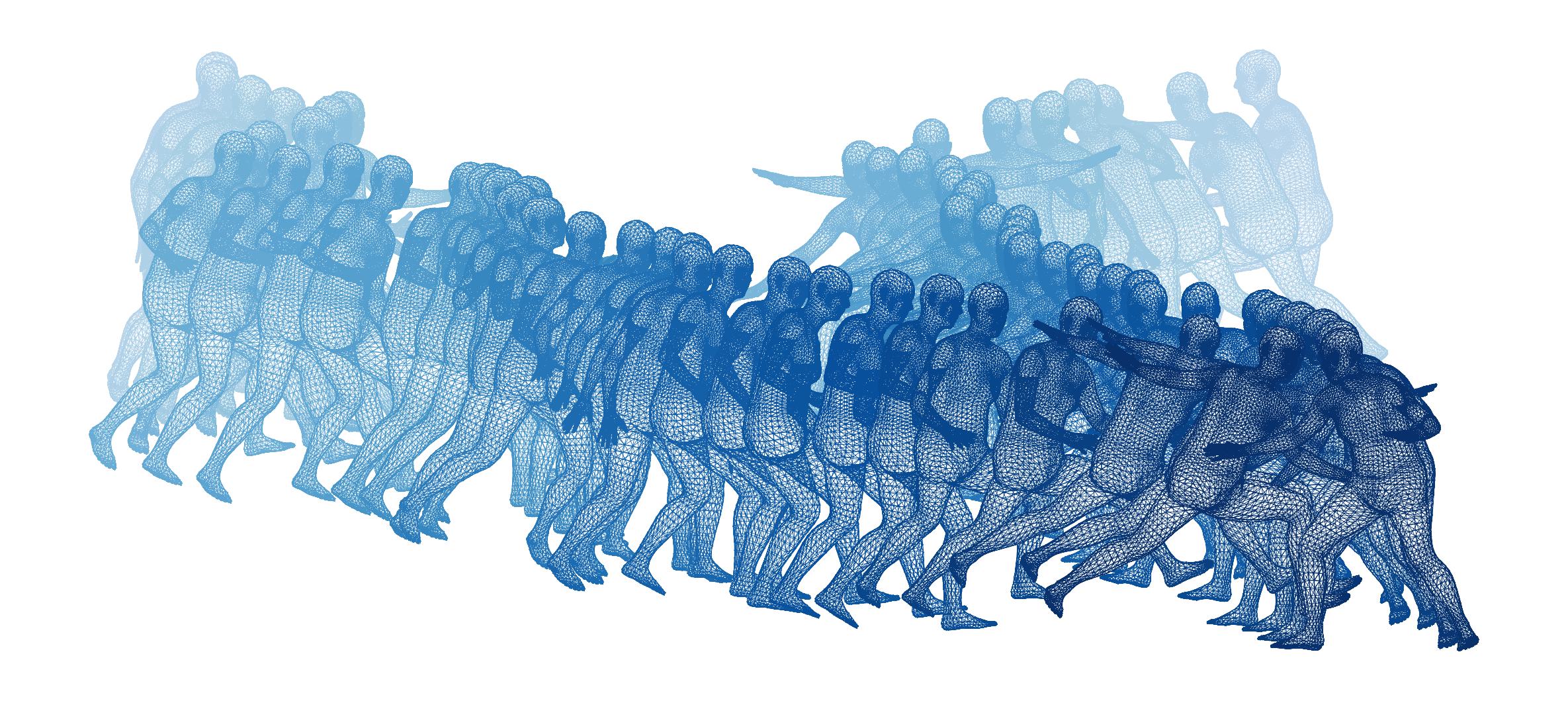} &
    \includegraphics[width=0.18\textwidth]{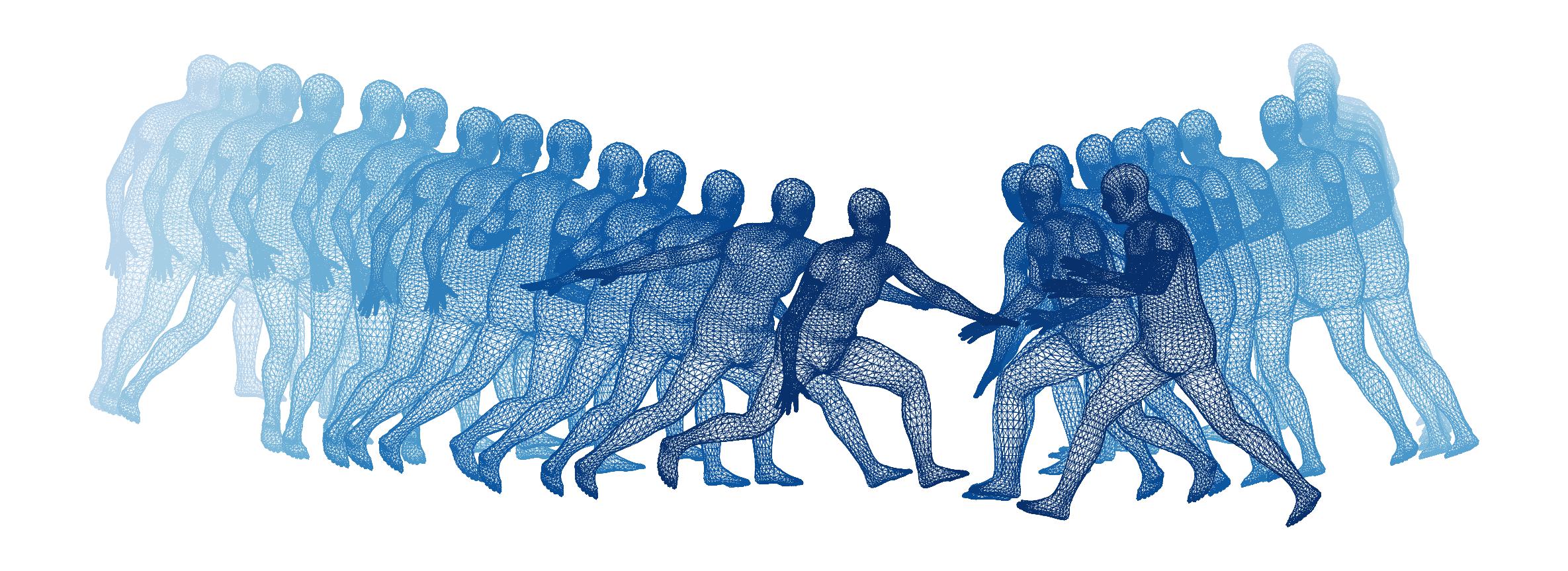} &
    \includegraphics[width=0.18\textwidth]{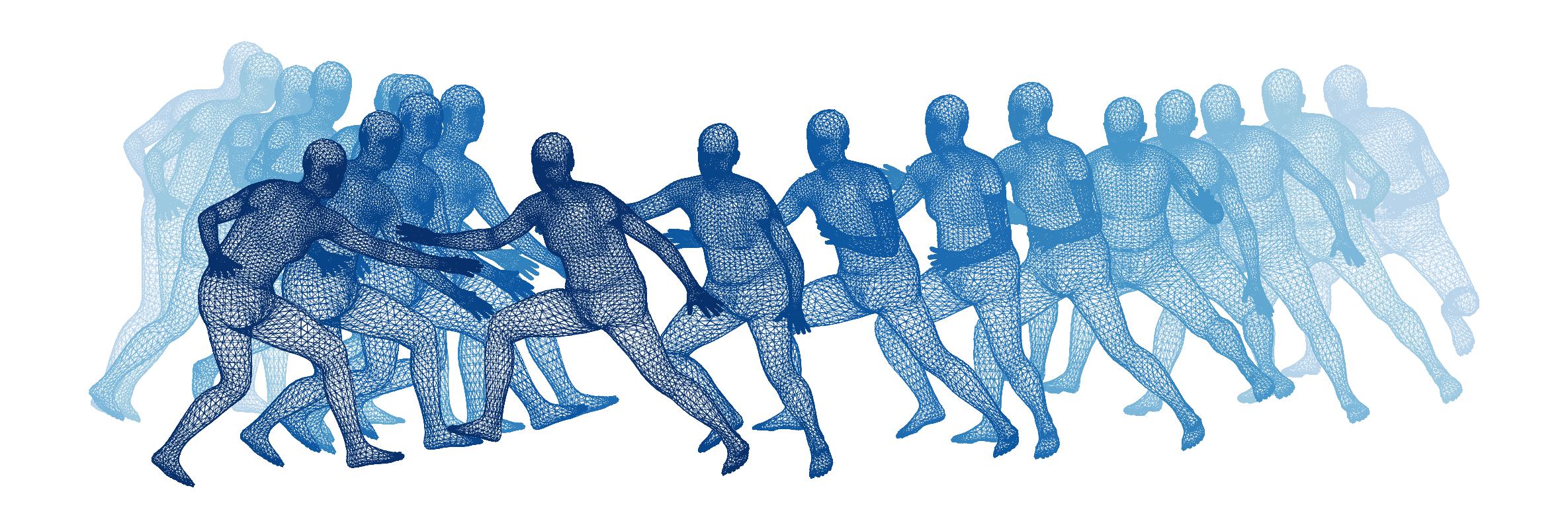} \\
    
    \includegraphics[width=0.18\textwidth]{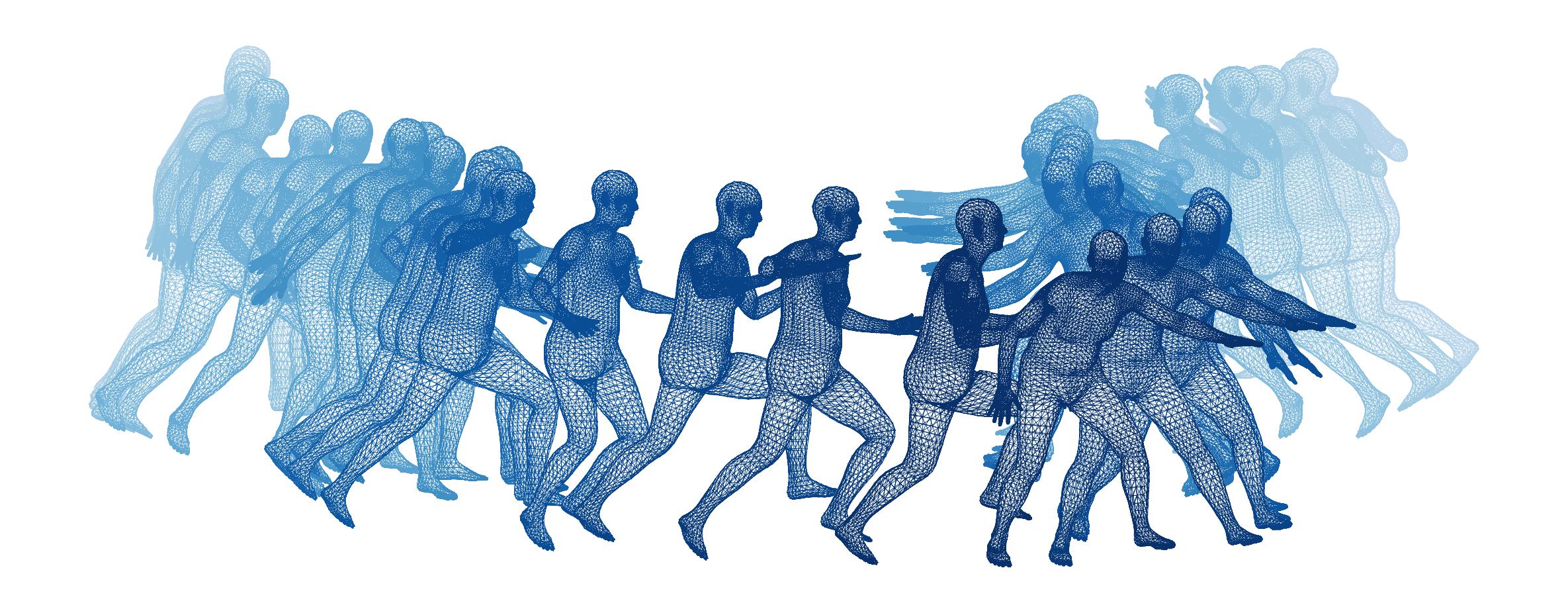} &
    \includegraphics[width=0.18\textwidth]{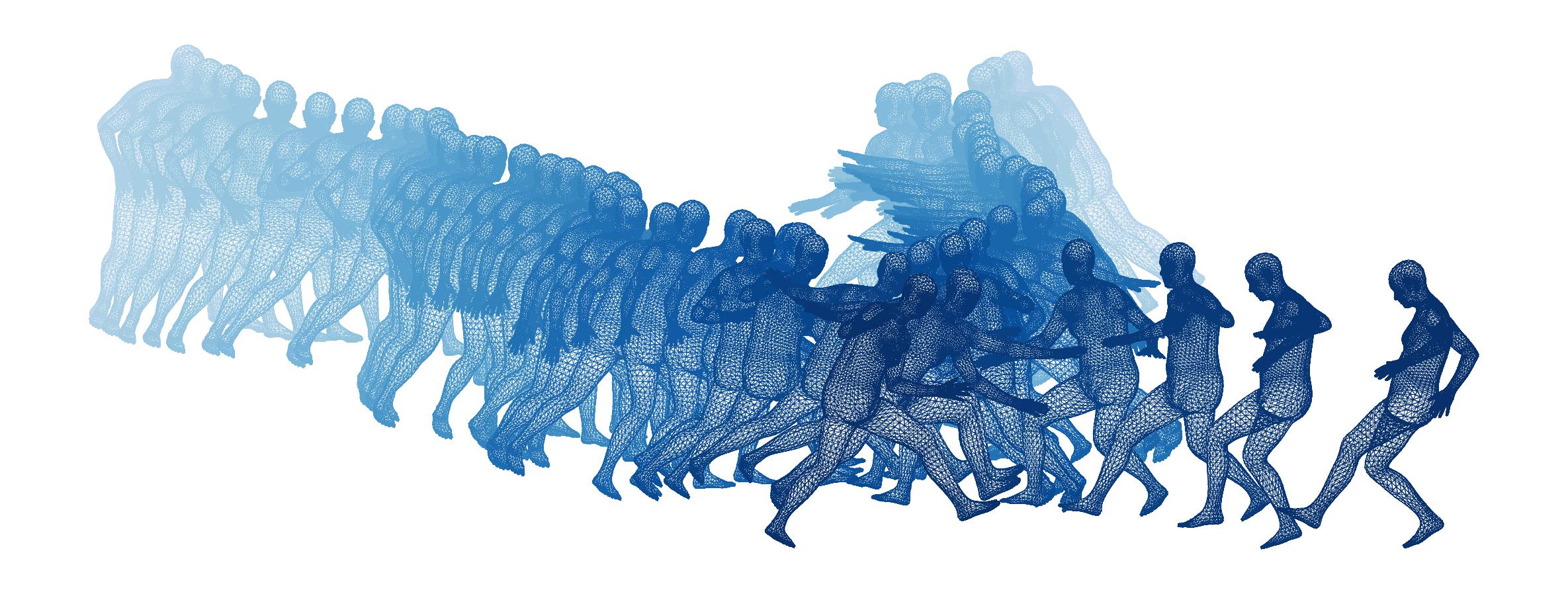} &
    \includegraphics[width=0.18\textwidth]{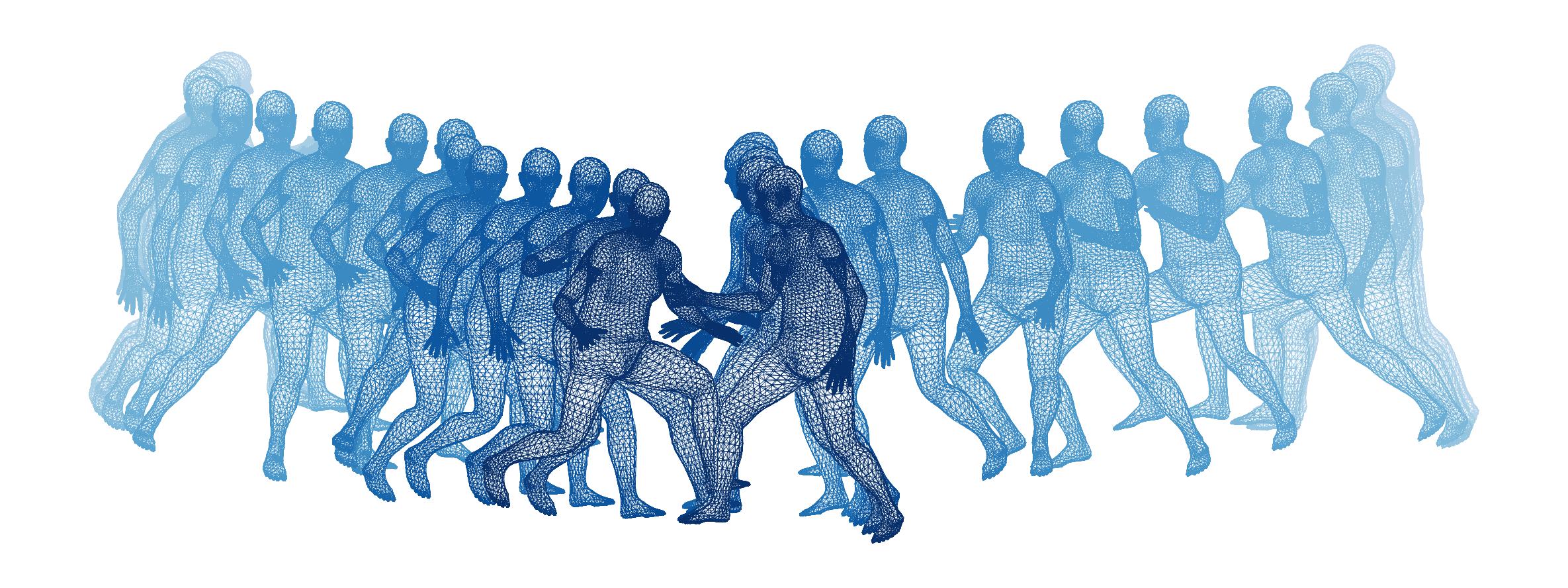} &
    \includegraphics[width=0.18\textwidth]{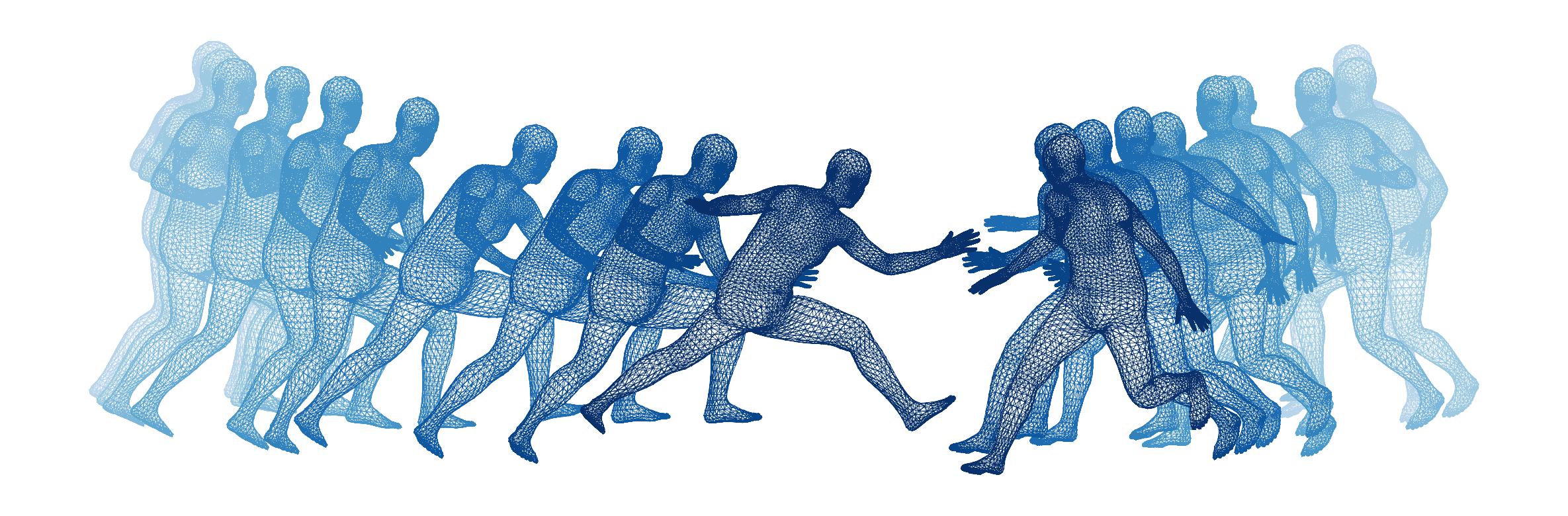} &
    \includegraphics[width=0.18\textwidth]{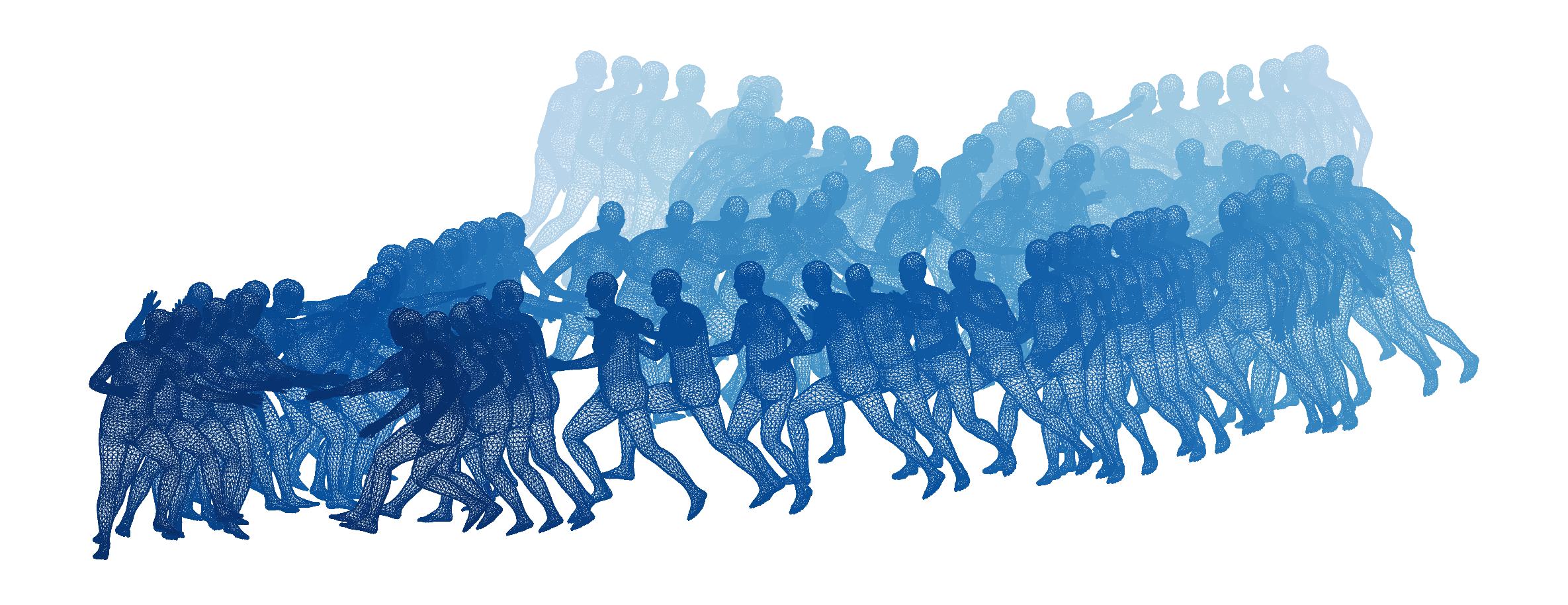} \\
    
    \includegraphics[width=0.18\textwidth]{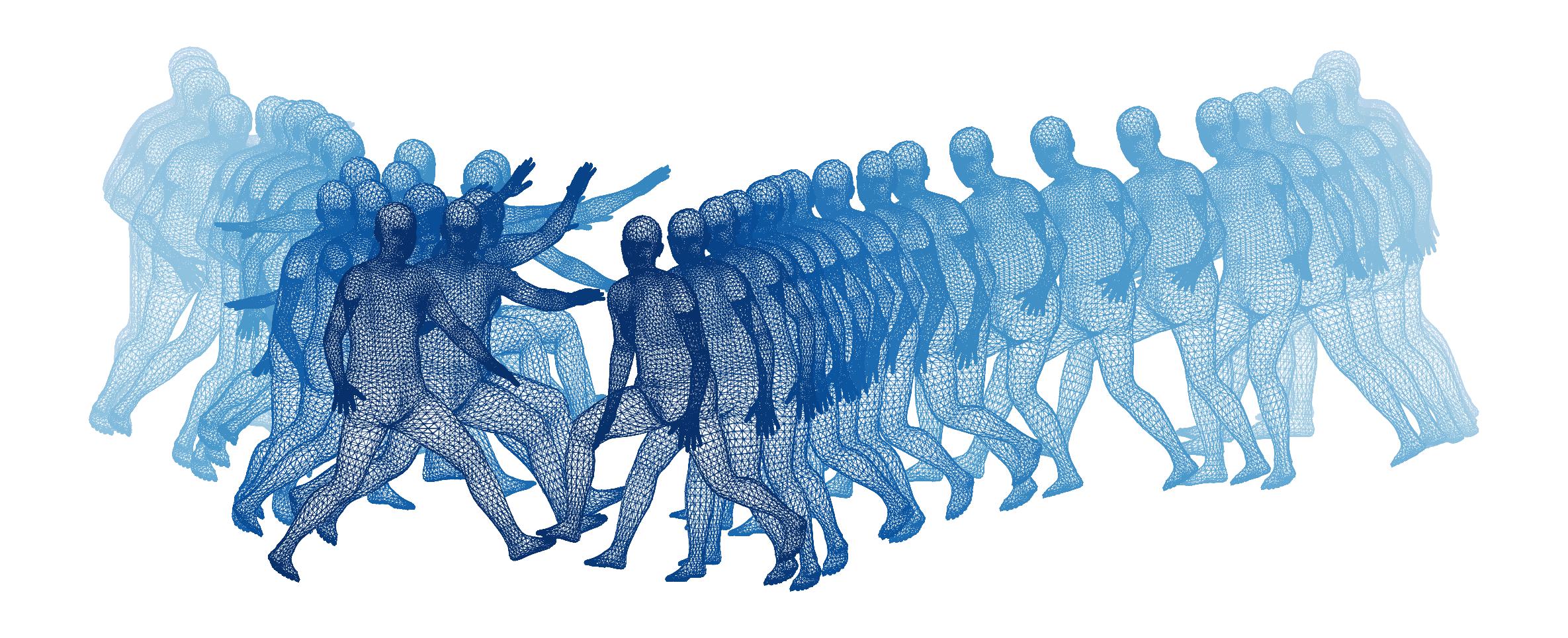} &
    \includegraphics[width=0.18\textwidth]{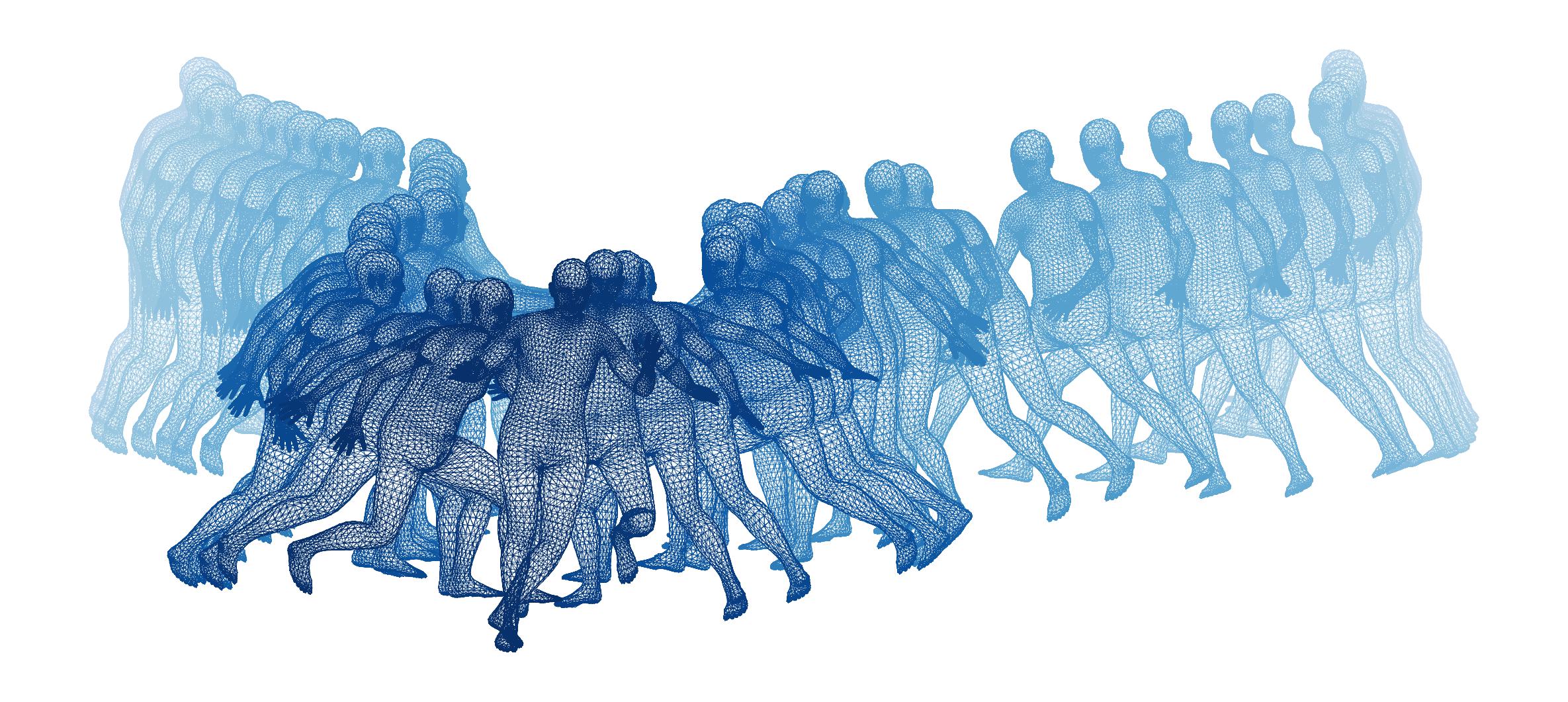} &
    \includegraphics[width=0.18\textwidth]{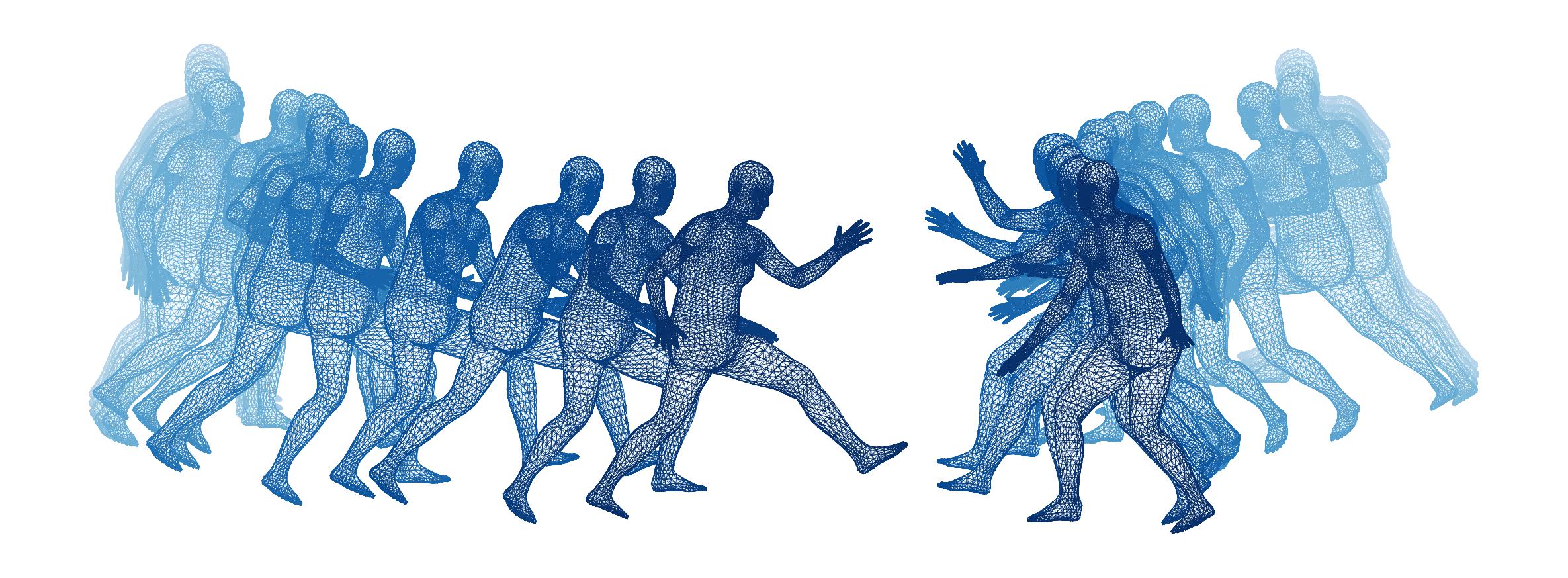} &
    \includegraphics[width=0.18\textwidth]{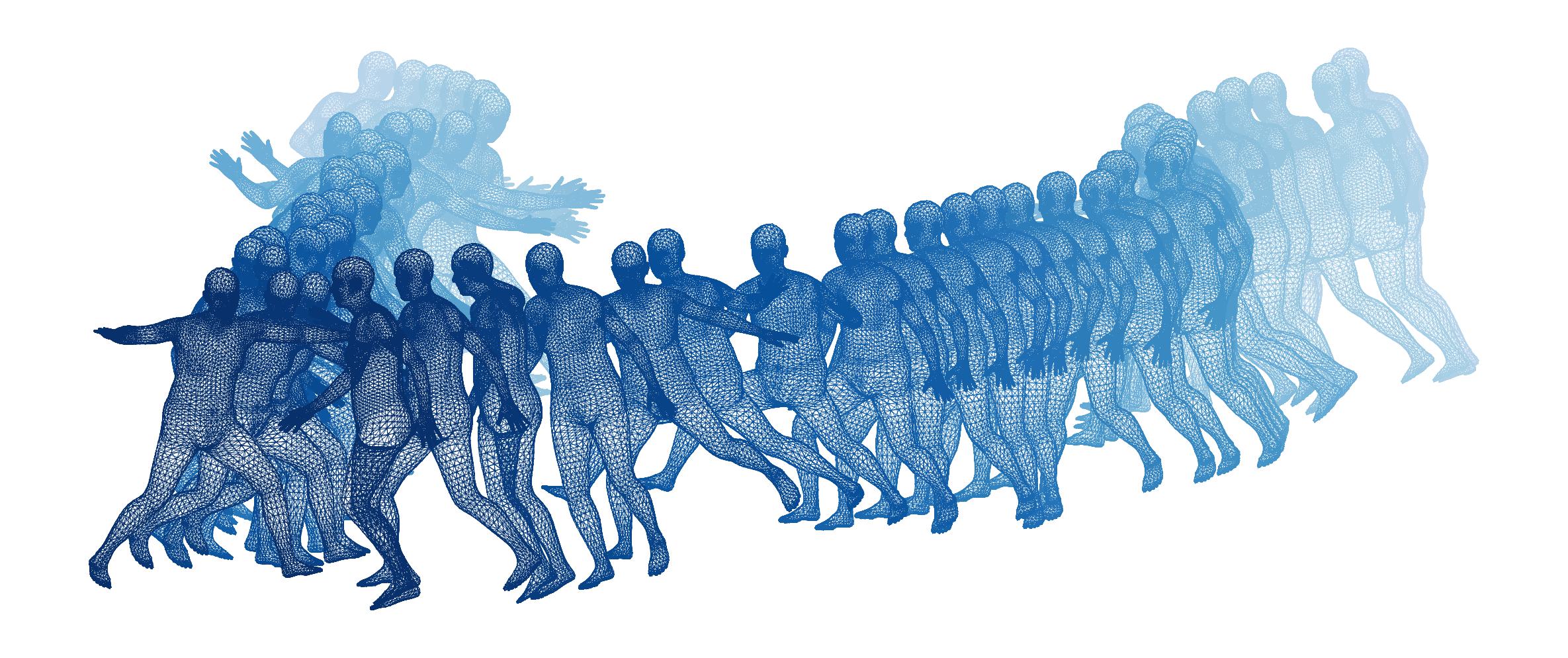} &
    \includegraphics[width=0.18\textwidth]{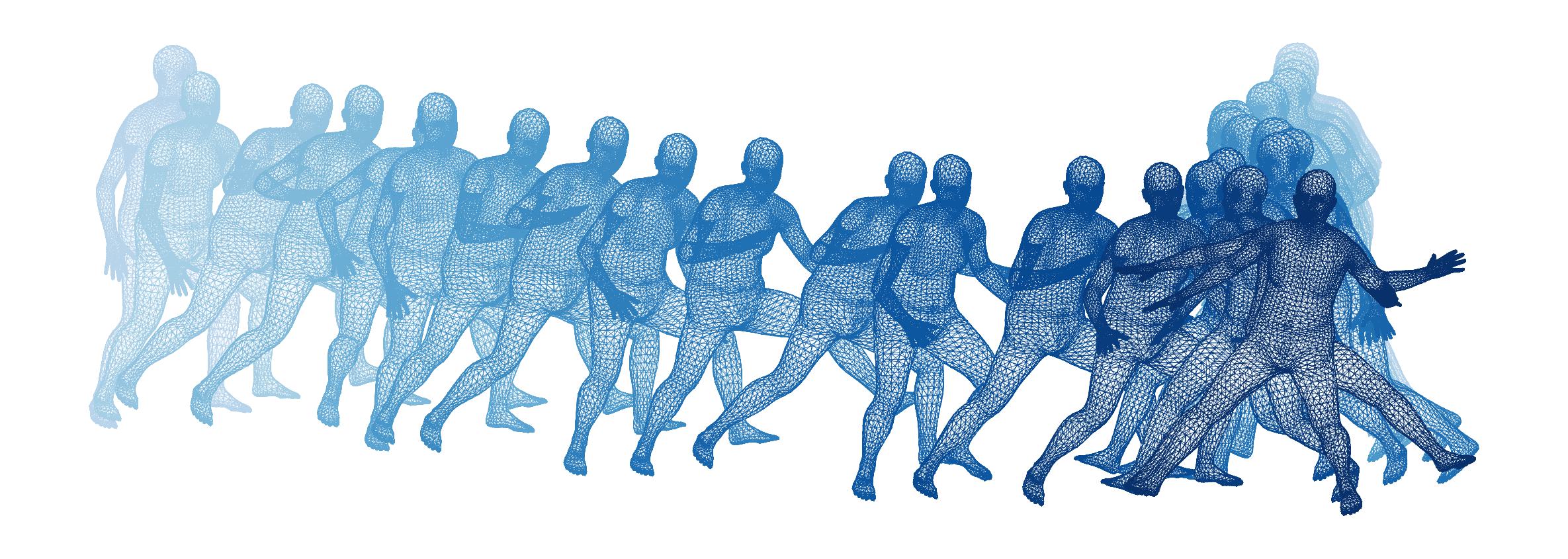} \\
    
    \includegraphics[width=0.18\textwidth]{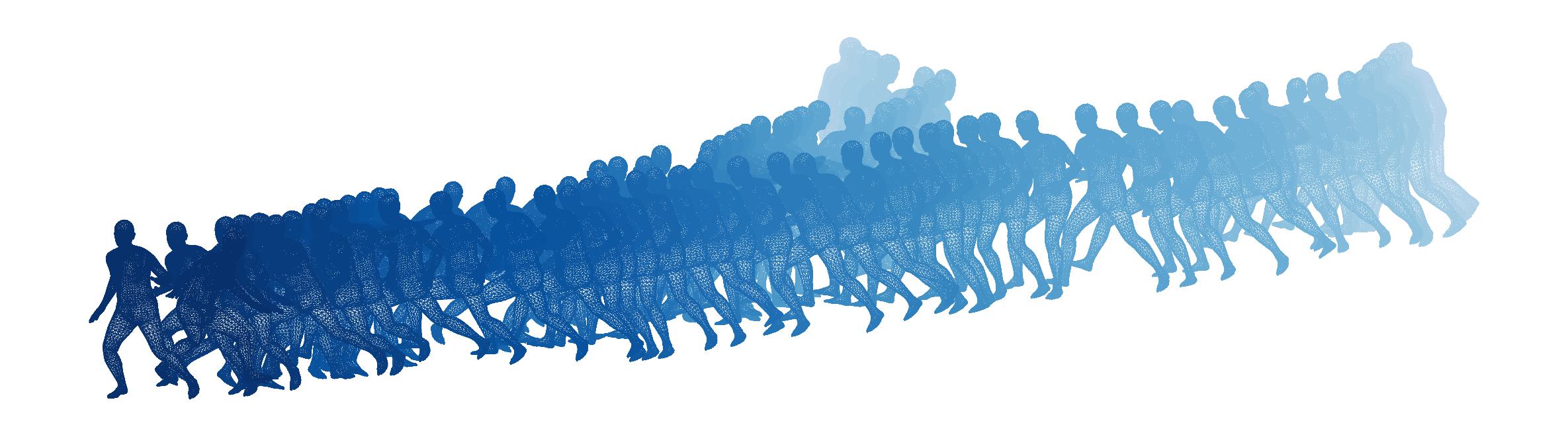} &
    \includegraphics[width=0.18\textwidth]{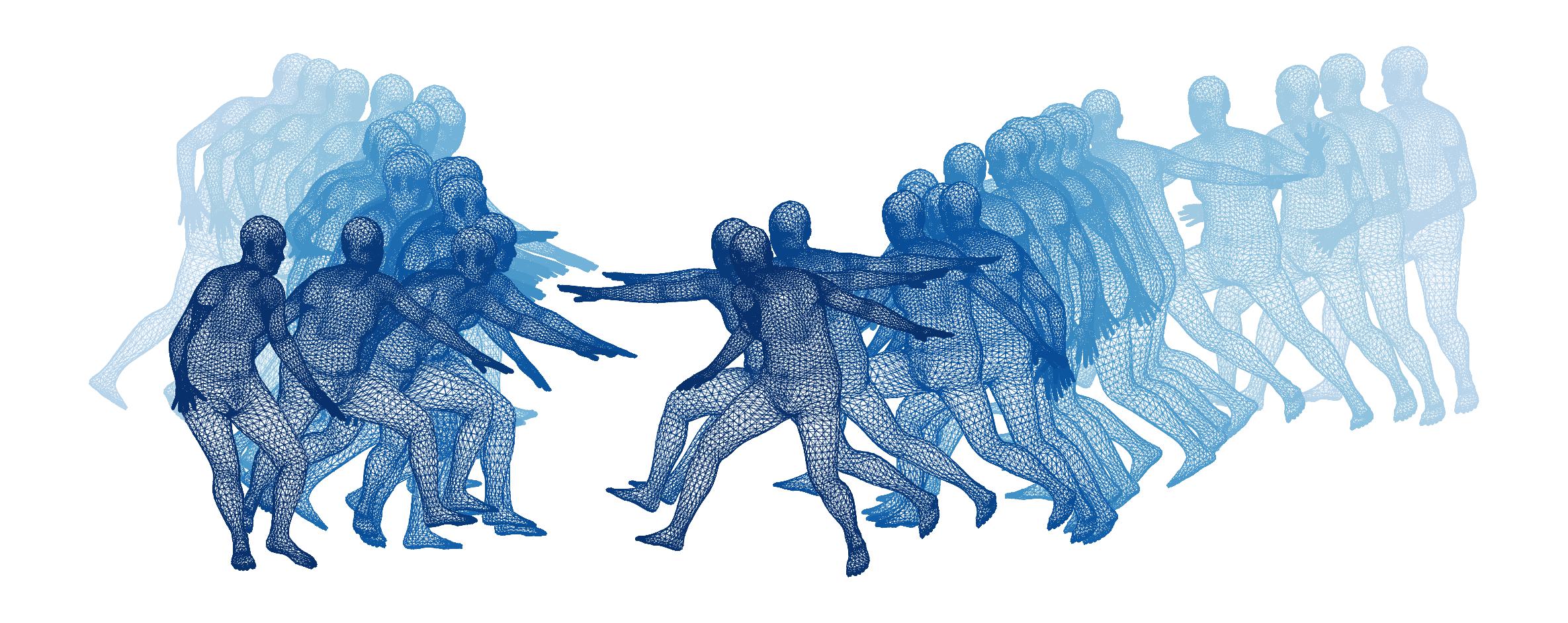} &
    \includegraphics[width=0.18\textwidth]{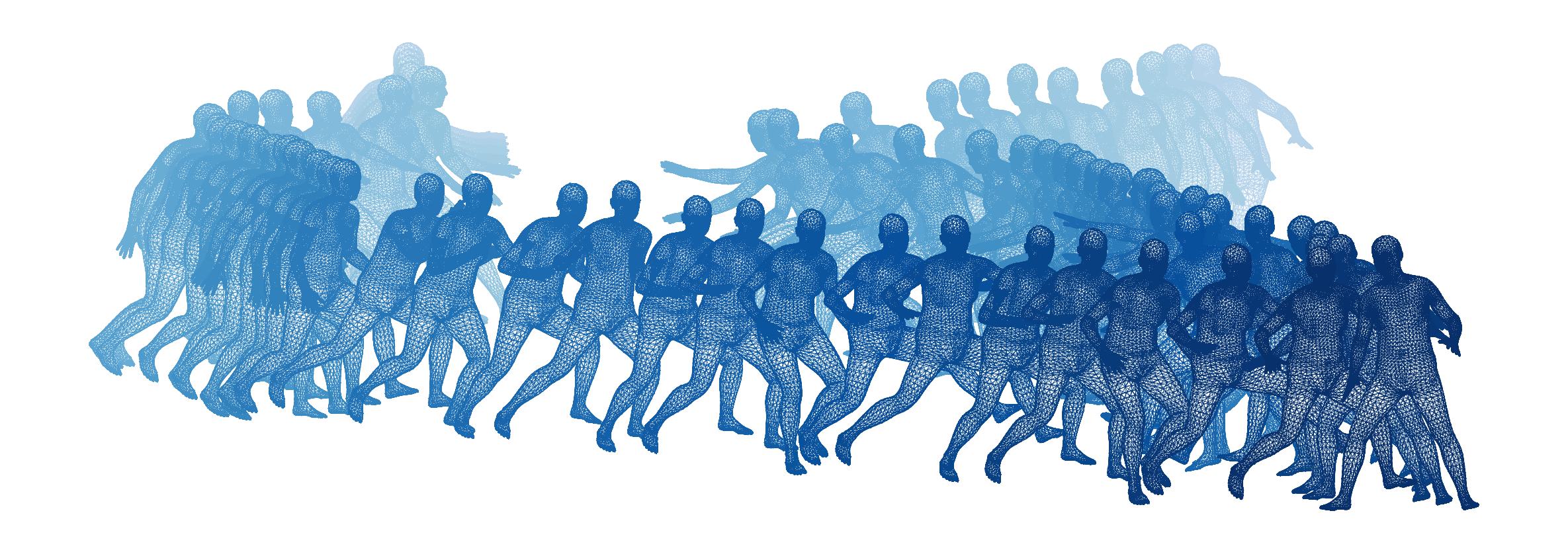} &
    \includegraphics[width=0.18\textwidth]{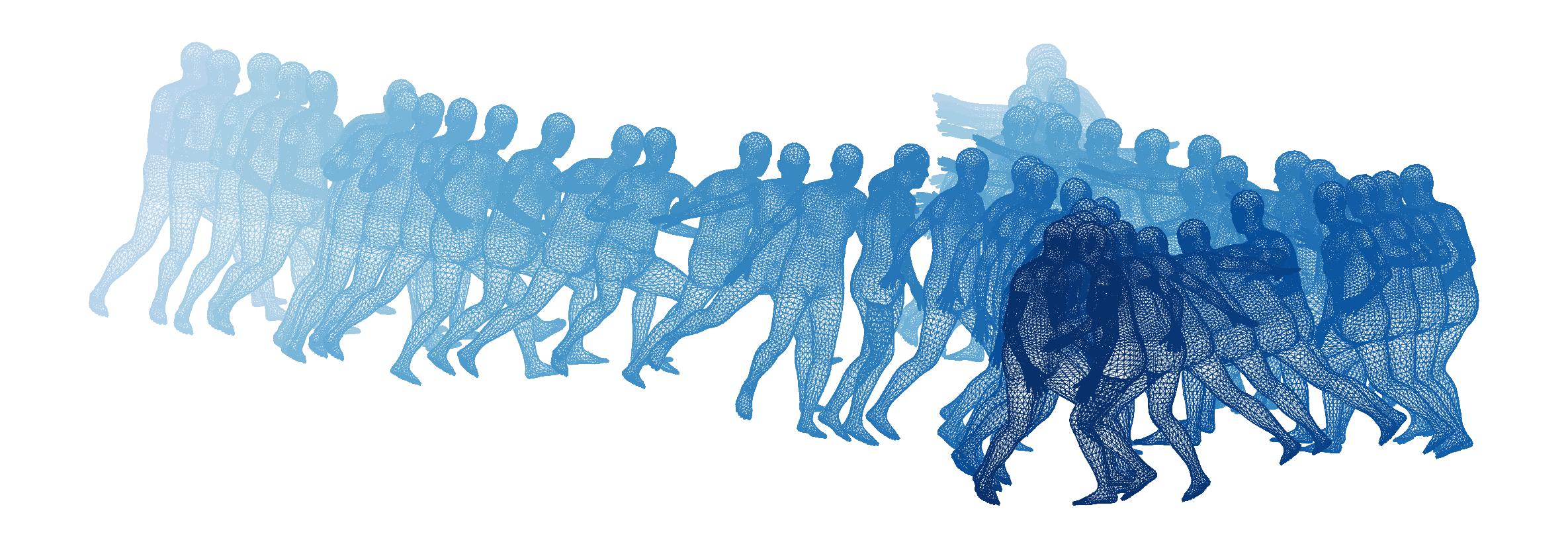} &
    \includegraphics[width=0.18\textwidth]{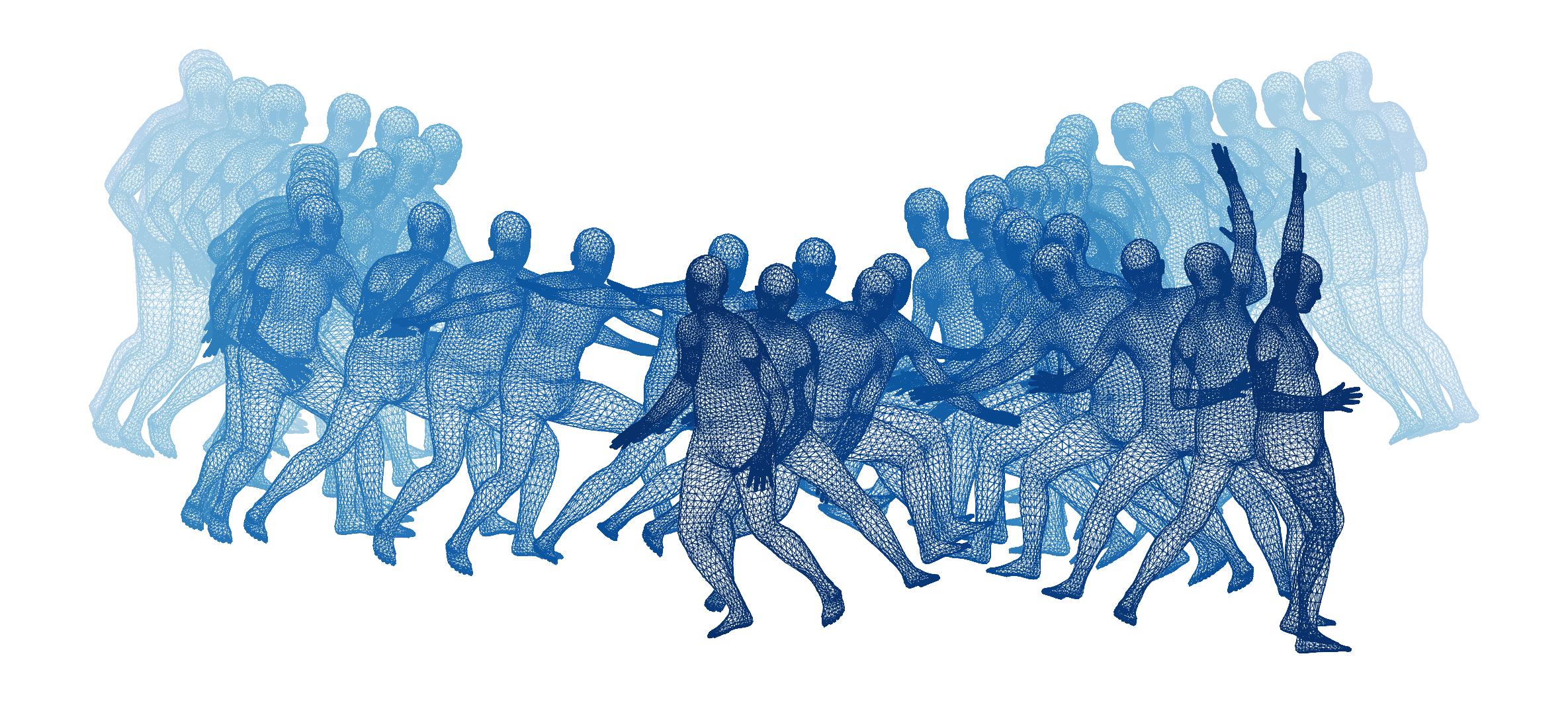} \\
  \end{tabular}
  \caption{Visualization of 25 overlay touches used in our user study of fencing against itself (self-play). Both left and right hand side motion sequences are generated from our strategy model. Top-to-bottom and light-to-dark indicate time progression.}
  \label{fig:user_study_vf_overlay_grid}
\end{figure*}

\clearpage
\newpage

\appendix

\section{Cluster labels}
\label{sec:cluster_labels}

\begin{table}[ht]
\centering
\caption{Cluster IDs, action labels, and per-cluster accuracy}
\begin{tabular}{clc}
\toprule
Cluster ID & Action Label & Accuracy \\
\midrule
C. 0  & Advance [large, chase down]                        & 0.9  \\
C. 1  & Off the line [massive advance]                     & 1.0  \\
C. 2  & Off the line [prep steps]                          & 0.9  \\
C. 3  & Retreat [medium]                                   & 0.8  \\
C. 4  & Off the line [medium advance, watching]            & 0.8  \\
C. 5  & Lunge [stop \& prep]                               & 1.0  \\
C. 6  & Advance [holding arm, chase down]                  & 1.0  \\
C. 7  & Provoke and Retreat                                & 0.8  \\
C. 8  & Retreat [arm out]                                  & 0.8  \\
C. 9  & React [short, parry, or lunge]                     & 0.7  \\
C. 10 & Parry/Close out                                    & 0.4  \\
C. 11 & Off the line [stutter steps]                       & 0.8  \\
C. 12 & Stop and Pull short                                & 0.7  \\
C. 13 & Advance [patient push, chase down]                 & 0.9  \\
C. 14 & Retreat [shuffle steps]                            & 0.7  \\
C. 15 & Retreat [counter attack]                           & 1.0  \\
C. 16 & Hit and Cheer                                      & 0.9  \\
C. 17 & Lunge [and turn to cheer]                          & 0.8  \\
C. 18 & Stop/Shift                                         & 0.8  \\
C. 19 & Off the line [check step]                          & 0.9  \\
C. 20 & Off the line [large step hop, watching]            & 0.9  \\
C. 21 & Advance [active arm]                               & 0.5  \\
C. 22 & Lunge [normal]                                     & 1.0  \\
C. 23 & Advance [medium, in the box]                       & 1.0  \\
C. 24 & Off the line [medium advance, aggressive]          & 0.9  \\
C. 25 & Advance [normal]                                   & 1.0  \\
C. 26 & Retreat [crossover]                                & 1.0  \\
C. 27 & Provoke and Pull Short                             & 1.0  \\
C. 28 & Stop cut                                           & 0.9  \\
C. 29 & Advance [balestra]                                 & 0.9  \\
\bottomrule
\end{tabular}
\label{tab:cluster_info}
\end{table}

\section{Expert feedback}
\label{sec:expert_feedback}

We collect several comments from the participants of the third user study in Table~\ref{tab:vf-feedback}.

\begin{table*}[ht]
\centering
\caption{Feedback from experts in User Study 3: \textit{Fence against professional fencers}}
\begin{tabular}{@{}p{0.95\linewidth}@{}}
\toprule
\textbf{Strengths} \\
\midrule
\textbullet\ "It understands right of way, and it knows when to attack and defend. This logic takes most beginning fencers many months to properly learn." \\[0.5em]
\textbullet\ "What stood out most to me was during the marching attack, VirtualFencer began taking a sequence of 'check steps' which implied to me that it understood the pacing of a long attack. Instead of 'chasing me down the strip' like a lot of beginner fencers often do, it took its time pushing me down the strip in stages, varying our distance as it did so as well. This was most impressive because this concept of 'pushing your opponent down the strip' is a more mature tactic."\\[0.5em]
\textbullet\ "VirtualFencer adjusted whenever I changed direction, seemed to take distance into account whenever an action was occurring, and was really good at decisions at the line (i.e. when I did C.19, it responded with C.18, which would make sense if the opponent was pulling short)."\\[0.5em]
\textbullet\ "It's quite good at recognizing when someone is moving forward, telling their pace or tempo, and adjusting based on the distance. It also appears to interpret speed well." \\[0.5em]
\textbullet\ "It started using more 'watching' tactics, either check stepping, or medium [watching] advances in the box. Because of this shift from decisive, faster actions to slower and more watching actions, it seems like it was strategically reacting to the way that I was choosing tactics." \\[0.5em]

\midrule
\textbf{Areas for improvement} \\
\midrule
\textbullet\ "There were instances where VirtualFencer had the right of way, yet chose to do actions that are typically done on the defense. For example, going forward with the marching attack but finishing with a stop cut or retreating for no reason." \\[0.5em]
\textbullet\ "Sometimes it crashes or takes unreasonably large/small steps based on the distance." \\[0.5em]
\textbullet\ "It does well with directionality and speed, but that's more about recognizing movement patterns rather than understanding the intent behind them. Strategy, even at a basic level, is theoretical—it's about the “why” behind a reaction, not just the reaction itself." \\[0.5em]
\textbullet\ "There's a layer of strategy that involves predicting and adapting based on your opponent's behavior. Right now, I don't think the system is capable of fully understanding those kinds of decisions—like why it's retreating, or why there's a reaction at all." \\[0.5em]
\textbullet\ "Compound footwork is also not taken into account (i.e., reprise, remises). This implies that for every one action your opponent takes, you will only take one action as well, whereas in reality, your opponent can take multiple steps in the time that it takes you to take one step." \\[0.5em]
\bottomrule
\end{tabular}
\label{tab:vf-feedback}
\end{table*}

\end{document}